\pdfoutput=1
\documentclass[11pt]{article}

\usepackage[final]{acl}
\hypersetup{colorlinks=true, linkcolor=blue, urlcolor=blue, citecolor=blue}

\usepackage{amsmath}
\usepackage{amssymb}
\usepackage{times}
\usepackage{latexsym}
\usepackage{booktabs}
\usepackage{float}
\usepackage{caption}
\usepackage{tabularx}
\usepackage{graphicx}

\usepackage{tcolorbox}  
\usepackage{lmodern}
\tcbuselibrary{skins,breakable}

\definecolor{promptbg}{RGB}{0,153,255}
\definecolor{jokebg}{RGB}{140,60,255}
\definecolor{cardborder}{RGB}{210,215,230}
\definecolor{slate}{RGB}{47, 62, 70}    
\definecolor{emerald}{RGB}{45, 106, 79}  
\definecolor{mintbg}{RGB}{240, 249, 246} 
\definecolor{ghostwhite}{RGB}{248, 249, 250}

\definecolor{p1color}{RGB}{0, 122, 204}      
\definecolor{p2color}{RGB}{180, 90, 0}       
\definecolor{p3color}{RGB}{40, 140, 70}      
\definecolor{p4color}{RGB}{140, 30, 140}     
\definecolor{p5color}{RGB}{190, 40, 40}      
\definecolor{p6color}{RGB}{30, 130, 160}     
\definecolor{jokegold}{RGB}{184, 134, 11}    
\definecolor{jokebg2}{RGB}{255, 252, 240}    

\newtcolorbox{personabox}[2][black]{
  enhanced,
  title={#2},
  colback=#1!5,
  colframe=#1!50,
  colbacktitle=#1,
  coltitle=white,
  fonttitle=\bfseries\sffamily\small,
  boxrule=0.5pt,
  leftrule=3pt,
  arc=3pt,
  titlerule=0pt,
  attach boxed title to top left={yshift=-2pt, xshift=6pt},
  boxed title style={
    colback=#1,
    colframe=#1,
    arc=2pt, boxrule=0pt,
    left=5pt, right=5pt,
    top=2pt, bottom=2pt
  },
  left=6pt, right=6pt, top=6pt, bottom=6pt,
  fontupper=\scriptsize\sffamily,
  bottom=4pt
}

\newtcolorbox{jokebox}{
  enhanced,
  colback=jokebg2,
  colframe=jokegold!70,
  leftrule=3pt,
  rightrule=0.4pt,
  toprule=0.4pt,
  bottomrule=0.4pt,
  arc=2pt,
  left=8pt, right=6pt, top=4pt, bottom=4pt,
  fontupper=\scriptsize\sffamily\itshape,
  before upper={{\color{jokegold}\bfseries\sffamily\scriptsize JOKE\quad}},
}

\definecolor{inputaccent}{RGB}{60, 60, 60}

\newtcolorbox{inputbox}{
  enhanced,
  colback=ghostwhite,
  colframe=inputaccent!40,
  boxrule=0.5pt,
  leftrule=3pt,
  arc=3pt,
  left=8pt, right=8pt, top=5pt, bottom=5pt,
  before upper={{\color{inputaccent}\bfseries\sffamily\scriptsize
    INPUT HEADLINE\quad}},
  fontupper=\small\sffamily\itshape,
}

\usepackage[T1]{fontenc}

\usepackage[utf8]{inputenc}

\usepackage{microtype}

\IfFileExists{inconsolata.sty}{\usepackage{inconsolata}}{}

\title{HumorGen: Cognitive Synergy for Humor Generation in Large Language Models via Persona-Based Distillation}

\author{
Edward Ajayi, Prasenjit Mitra \\
Carnegie Mellon University Africa \\
\texttt{\{eaajayi, prasenjm\}@andrew.cmu.edu}
}

\begin{document}
\maketitle

\begin{abstract}
Humor generation poses a significant challenge for Large Language Models (LLMs), because their standard training objective (next-token prediction) inherently conflicts with the surprise and incongruity required for comedy. To bridge this gap, we introduce the \textbf{Cognitive Synergy Framework}, a methodology for generating high-quality humor data inspired by psychological theories of humor. Utilizing a \textit{Mixture-of-Thought (MoT)} approach, we deploy six cognitive personas (e.g., \textit{The Absurdist}, \textit{The Cynic}) to synthesize diverse comedic perspectives for a given prompt. This framework produces a theory-grounded dataset, which we use to fine-tune a 7B-parameter student model. We further evaluate two alignment strategies, \textbf{Direct Preference Optimization (DPO)} and an offline group-relative variant \textbf{O-GRPO}, finding that neither improves over Supervised Fine-tuning (SFT). However, our 7B HumorGen model variants significantly outperform larger instruction-tuned baselines and achieve top-tier open-weight performance while remaining competitive with frontier proprietary systems. These results suggest that cognitively driven data curation is more critical than alignment algorithms or model scale for humor generation. 
\end{abstract}

\section{Introduction}

Humor generation is a sophisticated creative task requiring mastery of context, nuance, and linguistic ambiguity~\cite{khurana2024lolgorithm, robison2024lastlaugh}. While Large Language Models (LLMs) excel at logical reasoning, reliable humor generation remains an open problem because standard training objectives, which minimize perplexity, conflict with the incongruity and surprise required for comedy. \footnote{Project Website: \url{https://humorgen.pages.dev/}}
 This ``alignment tax'' often results in models that are safe and helpful but produce predictable, boring jokes or tedious explanations of humor. 

\begin{figure}[t]
\centering
\begin{tcolorbox}[
  enhanced,
  width=0.96\linewidth,
  colback=white,
  colframe=black!12,
  boxrule=0.8pt,
  arc=10pt,
  left=0pt, right=0pt, top=0pt, bottom=8pt,
  shadow={1pt}{-1pt}{0pt}{black!8}
]

  \begin{tcolorbox}[
    enhanced,
    colback=black!88,
    colframe=black!88,
    arc=8pt, sharp corners=south,
    boxrule=0pt,
    left=10pt, right=10pt, top=7pt, bottom=7pt
  ]
    \begin{tabular}{@{}l@{\hspace{6pt}}l@{}}
      {\footnotesize\color{black!35}\texttt{FIGURE -}} &
      {\small\color{white}\bfseries\sffamily LLM-Generated Joke}
    \end{tabular}
  \end{tcolorbox}

  \vspace{6pt}
  \hspace{8pt}\begin{minipage}{0.92\linewidth}

  {\scriptsize\sffamily\bfseries\color{promptbg} \textsc{Input Prompt: News Headline}} \\[-2pt]
  \begin{tcolorbox}[
    colback=promptbg!6,
    colframe=promptbg!25,
    arc=2pt,
    boxrule=0.5pt,
    leftrule=3pt,
    left=6pt, right=6pt, top=4pt, bottom=4pt,
    after skip=6pt
  ]
    \small\sffamily ``Denzel Washington reveals he doesn't watch movies anymore''
  \end{tcolorbox}

  {\scriptsize\sffamily\bfseries\color{jokebg} \textsc{LLM Generated Joke}} \\[-2pt]
  \begin{tcolorbox}[
    colback=jokebg!6,
    colframe=jokebg,
    leftrule=3pt,
    rightrule=0.4pt,
    toprule=0.4pt,
    bottomrule=0.4pt,
    arc=2pt,
    left=6pt, right=6pt, top=5pt, bottom=5pt,
    after skip=0pt
  ]
    \small\sffamily\itshape
    Denzel doesn't watch movies anymore. He just stares at his Oscar and
    whispers plot spoilers until the statue begs for a projector. Even his TV
    got jealous and started rehearsing monologues in the mirror, and now it's
    nominated for Best Supporting Flatscreen.
  \end{tcolorbox}

  \end{minipage}
  \vspace{2pt}

\end{tcolorbox}
\vspace{-8pt}
\caption{Example of an LLM-generated joke based on a news headline prompt, synthesized using the Cognitive Synergy Framework.}
\label{fig:headline_joke}
\end{figure}

Recent efforts to improve LLM humor generation have focused on logical ``thought leaps''~\cite{zhong_lets_2024} or multistep reasoning~\cite{wang_innovative_2025}. While these approaches improve performance on specific humor generation tasks, they often miss the diverse cognitive styles underlying human humor, and primarily rely on instruction tuning, limiting their ability to capture the variety of ways humans construct jokes.

To bridge this gap, we introduce the \textbf{Cognitive Synergy Framework} which advance beyond generic instruction tuning by operationalizing psychological humor theories within a \textit{Mixture-of-Thought (MoT)} framework designed to encourage creative divergence. Traditional language modeling is highly susceptible to mode collapse in creative generation, converging toward the most probable (and therefore most generic) continuations. By instantiating six distinct ``cognitive personas'' (e.g., \textit{The Absurdist}, \textit{The Cynic}) as latent experts within the MoT framework, we consistently route the generation process into the low-probability, high-variance regions of the semantic space where humor naturally occurs. This ensemble approach mitigates mode collapse and yields a diverse, theoretically grounded corpus of synthetic data, enabling us to distill multi-faceted humor generation capabilities from a frontier teacher model into a highly efficient 7B-parameter student model.

Due to the highly subjective nature of humor, we investigate whether preference alignment (e.g., \textbf{Direct Preference Optimization (DPO)} and \textbf{Offline Group Relative Policy Optimization (O-GRPO)}) improves over supervised fine-tuning. Our experiments show that neither alignment method visibly improves the models over the SFT baseline as DPO achieves similar performance to SFT, while O-GRPO is less impressive. Thus, under our setup, the alignment exercises did not improve the models, and the quality of the underlying cognitive data (Cognitive Synergy Framework) is the primary driver of generation performance.

Our contributions are:
\begin{itemize}
    \item We introduce the \textbf{Cognitive Synergy Framework (CSF)}, a methodology that operationalizes psychological humor theories into six cognitive personas to generate diverse, theory-grounded humor data via Mixture-of-Thought.
    \item We investigate whether preference alignment (DPO, O-GRPO) improves over SFT for humor generation, finding that neither method yields consistent gains beyond a well-curated SFT baseline in this subjective humor generation domain.
    \item We introduce \textbf{Humor Transfer Bench (HTB)}, a 400-prompt benchmark spanning eight input domains to test generalization in textual humor generation.
    \item We show that our 7B student models, \textbf{HumorGen} variants, achieves top-tier open-weight performance on HTB and SemEval MWAHAHA humor-generation benchmark and remains competitive with much larger proprietary systems, showing that high-quality data can outweigh model size for humor.
\end{itemize}

\section{Related Work}

\subsection{Computational Humor Generation}


Computational humor research has largely focused on detection and recognition tasks~\cite{jentzsch2023chatgpt, dsilva2024augmenting}, while humor generation remains comparatively underexplored. As a result, existing work on humor generation is often fragmented, targeting specific forms such as puns~\cite{chen_are_2024}, domain-based humor~\cite{shafiei_not_2025}, punchline generation~\cite{zhang2020let}, or language-based humor generation~\cite{chen2023can, zhong_lets_2024}. While classical humor theories~\cite{lintott2016superiority, scheel2025definitions, mcgraw2010benign, ajayiautomatic} offer valuable linguistic and semantic interpretations of humor, they traditionally fail to translate into reliable generative mechanisms. To bridge this gap, our work directly operationalizes these psychological theories into generative mechanisms by encoding them into distinct ``Cognitive Personas'', allowing language models to explicitly construct humor based on established theoretical foundations.

\subsection{Reasoning-Enhanced Humor Generation}

With the rise of Large Language Models (LLMs), recent work has explored prompting and reasoning strategies for humor generation. Prior studies show that Chain-of-Thought (CoT) reasoning~\cite{wei2022chain}, while effective for logical tasks, is often misaligned with humor generation, which depends on divergence, incongruity, and non-linear associations~\cite{zhong_lets_2024, wang_innovative_2025, tikhonov_humor_2024}. As a result, CoT-based prompting often produces coherent but non-comedic outputs.


To address this, \textit{Creative Leap of Thought (CLoT)}~\cite{zhong_lets_2024} introduces leap-of-thought reasoning to encourage non-obvious associations during humor generation. The \textit{LoL} framework~\cite{wang_innovative_2025} builds on \textit{CLoT} by incorporating external knowledge to support multi-hop creative reasoning, while related work explores structured multistep reasoning for humor generation~\cite{tikhonov_humor_2024}. In contrast, naive prompting with ChatGPT~\cite{jentzsch2023chatgpt} shows strong repetition patterns, with 90\% of 1,008 generated jokes repeating only 25 templates, suggesting that standard reasoning paradigms remain limited for humor generation. Unlike these approaches that attempt to optimize a single, modified reasoning path, we propose using a Mixture-of-Thought (MoT) paradigm to run multiple, divergent reasoning traces in parallel, fundamentally shifting from searching for one ``correct'' logical path to generating a broad, creative candidate pool.

\subsection{Preference Optimization in Creative Generation}

Preference optimization methods have been widely studied for aligning LLMs with human preferences in subjective generation tasks~\cite{yasuda2025automatic, lou2025sequential, vikhorev2024cleancomedy}. With methods such as Direct Preference Optimization (DPO) and Group Relative Policy Optimization (GRPO) having demonstrated effectiveness across domains such as code and image generation~\cite{govandeteaching, tong2025delving}.


In humor generation, prior work integrates Supervised Fine-Tuning (SFT) with DPO-based alignment~\cite{wang_innovative_2025}. However, many preference optimization methods rely on online sampling or pairwise comparisons, which can be unstable and computationally expensive in highly subjective settings like comedy. To overcome this instability, we introduce an offline group-relative formulation (O-GRPO) that leverages a static dataset of diverse candidates, providing a more stable and computationally efficient alignment signal specifically tailored for subjective, creative generation tasks.



\section{The Cognitive Synergy Framework}

Humor generation needs both coherence and surprise: a joke must first establish a plausible setup, then introduce an unexpected shift. Standard LLM decoding tends to favor high-probability next-token predictions, which often reduces surprise and yields generic outputs. 

To address this, we introduce the \textbf{Cognitive Synergy Framework (CSF)}, which leverages Mixture-of-Thought (MoT)~\cite{fein2025mixture} for humor generation by running multiple persona-guided reasoning paths in parallel, where each \textit{Cognitive Persona} (see Section~\ref{cog_personas}) encodes a different humor mechanism (e.g., absurdity, social critique, wordplay), enabling the model to explore diverse comedic directions instead of converging to a single safe continuation. The resulting candidate set is therefore both coherent and diverse, improving downstream selection and distillation for humor-focused training.

\subsection{Divergent Reasoning via MoT}

Unlike standard Chain-of-Thought (CoT) prompting, which optimizes for a single logical path, our framework generates $K$ distinct reasoning traces in parallel. Given an input premise $x$, we sample a set of diverse reasoning paths $\{z_1, z_2, \dots, z_K\}$ seeded by different cognitive priors, yielding a \textit{pool} of $K$ candidates that is retained in full. All candidates are subsequently ranked and used for training, so the model learns from the full distribution of generated joke candidates rather than a single top-ranked joke.

\subsection{Cognitive Personas}
\label{cog_personas}
To guide this diversity, we define six \textbf{Cognitive Personas}, each grounded in a specific psychological theory of humor (Table~\ref{tab:personas}; see Appendix~\ref{sec:appendix_humor_theories} for a full overview of the underlying theories). These personas act as soft constraints on the reasoning process, ensuring that our candidate pool covers a wide spectrum of comedic mechanisms.

\begin{figure*}[t]
    \centering
    \includegraphics[width=\textwidth]{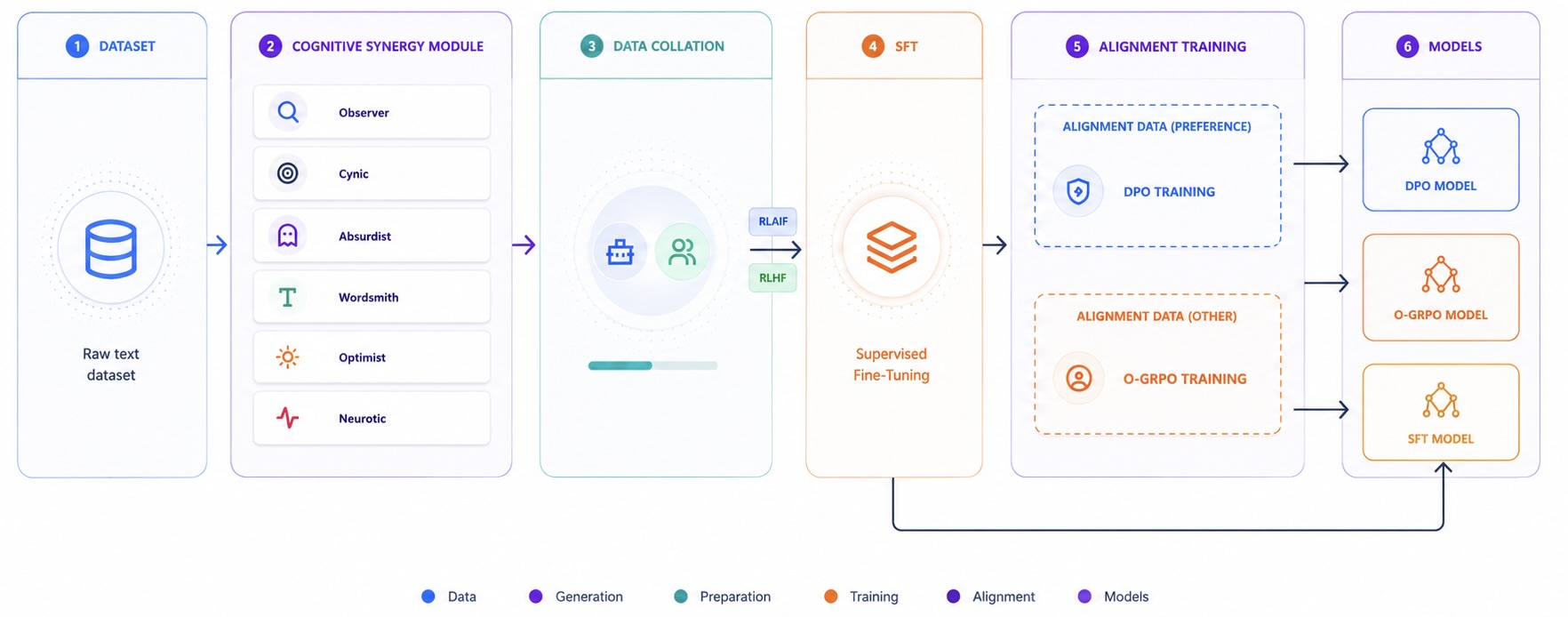}

    \caption{The HumorGen training pipeline. Input headlines are processed by the Cognitive Synergy module (MoT), generating diverse candidates from 6 distinct personas. Candidates are ranked via pairwise LLM judging (Llama 3.3-70B) to compute Elo ratings. The base policy is fine-tuned on the top-ranked candidates. The model is further optimized via two parallel experimental branches: Pairwise DPO (top) or Group-Relative O-GRPO (bottom) driven by Elo-based preference data.}
    \label{fig:architecture}
\end{figure*}

\begin{table*}[t]
\centering
\small
\begin{tabularx}{\textwidth}{@{}l l l X@{}}
\toprule
\textbf{Persona} & \textbf{Humor Theory} & \textbf{Mechanism} & \textbf{Cognitive Focus} \\ 
\midrule
\textbf{Neurotic} & Relief Theory & Tension Release & Internal anxiety, overthinking, and social insecurity. \\ 
\textbf{Cynic} & Superiority Theory & Social Critique & Hypocrisy, biting sarcasm, and moral contradictions. \\ 
\textbf{Observer} & Incongruity & Social Mapping & Mundane minutiae and unwritten awkward social norms. \\ 
\textbf{Wordsmith} & Linguistic & Ambiguity & Puns, double entendres, and phonological play. \\ 
\textbf{Optimist} & Benign Violation & Recontextualization & Wholesome misinterpretations of potentially negative traits. \\ 
\textbf{Absurdist} & Incongruity & Surrealism & Non-sequiturs, dream logic, and fractured causality. \\ 
\bottomrule
\end{tabularx}
\caption{The six Cognitive Personas used in our framework. We map each persona to a foundational humor theory and a specific cognitive focus to ensure divergent candidate generation.}
\label{tab:personas}
\end{table*}

By using these personas, we created a ``synergy'' between different styles of thought. This structural diversity proved critical at the supervised fine-tuning stage, providing a rich variety of distinct candidates from which the student model could learn humor generation.

\section{Methodology}

We frame humor generation as a conditional language modeling task, where the goal is to generate a humorous response $y$ given a context $x$. We first build a strong initial model via supervised fine-tuning (SFT) using MoT-generated joke candidates and LLM-ranked supervision. Starting from this model, we explore two preference alignment strategies based on LLM-judged pairwise evaluations: Direct Preference Optimization (DPO)~\cite{rafailov2023direct} and an offline group-relative variant, \textbf{O-GRPO}, both of which aim to better align outputs with learned preference signals.

\subsection{Supervised Fine-Tuning (SFT)}
This initial stage establishes baseline humor capabilities and internalizes the various cognitive personas. We construct a dataset $\mathcal{D}_{SFT}$ using an ``LLM-ranking'' protocol. Given the candidate pool $\mathcal{C}_{total}$ generated by our Mixture-of-Thought (MoT) ensemble, we employ a pairwise LLM evaluation system to compute Elo ratings for all candidates. We select the top-ranked candidates for each prompt based on these Elo ratings:
$$ y^* = \text{argmax}_{y \in \mathcal{C}_{total}} \text{Score}_{LLM}(y|x) $$
We fine-tune the student model using standard cross-entropy loss to maximize the likelihood of these ``winner'' responses:
$$ \mathcal{L}_{SFT}(\theta) = -\mathbb{E}_{(x, y^*) \sim \mathcal{D}_{SFT}} [ \log \pi_\theta(y^* | x) ] $$
This stage effectively distills the creative diversity of the larger teacher model into the student model.

\subsection{Direct Preference Optimization (DPO)}
To further align the model with humor preferences, we employ DPO using a dataset $\mathcal{D}_{DPO}$ of high-quality pairwise preferences derived from the LLM-judged Elo rankings. Each pair $(y_w, y_l)$ consists of a high-ranking joke $y_w$ and a low-ranking candidate $y_l$ for the same prompt, selected based on their Elo gap. We optimize the policy $\pi_\theta$ directly without a reward model:

\begin{multline}
\mathcal{L}_{DPO}(\pi_\theta; \pi_{ref}) = - \mathbb{E}_{(x, y_w, y_l) \sim \mathcal{D}_{DPO}} \Bigg[ \log \sigma \Bigg( \\
\beta \log \frac{\pi_\theta(y_w|x)}{\pi_{ref}(y_w|x)} - \beta \log \frac{\pi_\theta(y_l|x)}{\pi_{ref}(y_l|x)} \Bigg) \Bigg]
\end{multline}

\subsection{Offline Group Relative Policy Optimization (O-GRPO)}
Beyond pairwise preference alignment, we experiment with an offline variant of Group Relative Policy Optimization, \textbf{O-GRPO}, which adapts GRPO~\cite{shao2024deepseekmath} to a fixed preference dataset. Rather than sampling candidates and computing rewards online during training, O-GRPO operates on the pre-constructed candidate pool: all generated joke candidates per prompt are ranked once via a complete Bradley-Terry tournament, and group-relative advantages are derived from those static scores. Specifically, for each joke $y_i$, we compute its advantage as
\begin{equation}
A_i = \frac{r_i - \mu_{\mathrm{group}}}{\sigma_{\mathrm{group}} + \epsilon}
\label{eq:ogrpo_main}
\end{equation}
where $r_i$ is its pre-computed ranking score. We then apply exponentially weighted SFT, where each candidate contributes proportionally to the weights defined in Appendix~\ref{sec:appendix_ogrpo}. This upweights relatively stronger responses and downweights weaker ones using fixed offline rankings, without live reward computation.





\subsection{Cognitive Synergy Distillation (CSD)}
\label{sec:csd}
In the SFT stage of the training pipeline, the teacher's persona-specific reasoning traces are not used, and the student model observes only the final jokes. In contrast, CSD trains the student on both the reasoning traces and the corresponding joke:

\noindent\texttt{<think>} \textit{persona-specific brainstorming} \texttt{</think>} \textit{joke}

This is process distillation: the student learns not just \textit{what} to generate but \textit{how} the teacher planned it. For DPO, both chosen and rejected responses include reasoning traces (symmetric format), so the model cannot shortcut by learning that the mere presence of reasoning correlates with winning; it must learn which \textit{content} leads to better jokes.

At inference, the model generates reasoning followed by the joke. The reasoning is stripped for evaluation (ensuring fair comparison with non-CSD models) but retained for interpretability. Unlike generic CoT, ineffective for humor~\cite{zhong_lets_2024, tikhonov_humor_2024}, CSD's reasoning is grounded in specific humor theories through the cognitive personas, making it a form of \textit{theory-grounded creative distillation}.

\section{Experimental Setup}

\subsection{Datasets and Data Synthesis}
We utilize the official SemEval 2026 Task 1 (MWAHAHA) experimental set~\cite{semeval2026mwahaha}, comprising 1,200 news headlines and word-pair prompts as inputs to our generation pipeline. Using the Cognitive Synergy Framework, we generate 24 candidates per prompt (4 per persona $\times$ 6 personas) from a teacher ensemble of \textit{Kimi-K2} and \textit{Qwen 2.5-32B-Instruct}, yielding a raw pool of $\sim$28,800 candidates. Details on the teacher model selection process are provided in Appendix~\ref{sec:appendix_initial_models}. These candidates are scored and ranked via a pairwise LLM evaluation system using Llama 3.3-70B-Instruct as the judge, producing per-prompt Elo ratings for all 24 candidates. Details of the judge selection process are provided in Appendix~\ref{sec:appendix_judge_selection}. We construct three training subsets from these rankings:

\begin{itemize}
    \item SFT Data ($\mathcal{D}_{SFT}$, $N=12,000$): For each of the 1,200 prompts, we select the top 10 Elo-ranked candidates (rather than only the single best). Using multiple top-ranked candidates per prompt avoids mode collapse: the student learns a diverse range of humor styles (e.g., wordplay, absurdity, sarcasm) instead of collapsing toward one dominant style.
    \item DPO Data ($\mathcal{D}_{DPO}$, $N=6,000$): For each prompt, we construct 5 preference pairs by randomly pairing candidates from the top-5 Elo-ranked jokes (chosen, $y_w$) with candidates from the bottom-5 Elo-ranked jokes (rejected, $y_l$). This yields 5 pairs $\times$ 1,200 prompts = 6,000 preference pairs, with a sharp quality gap between chosen and rejected responses.
    \item O-GRPO Data ($\mathcal{D}_{GRPO}$): We use all 24 candidates per prompt across the 1,200 prompts, computing z-score group-relative Elo advantages per group ($G=24$). This exposes the model to the full quality spectrum within each prompt group.
\end{itemize}
Appendix~\ref{sec:appendix_training_distribution} reports the persona composition of each of the training and alignment corpus, confirming that all six cognitive personas contribute to SFT, DPO, and O-GRPO training data.


For evaluation, we use two benchmarks. (1) \textbf{SemEval-MWAHAHA}: from the official 300-prompt test set~\cite{semeval2026mwahaha}, we evaluate on the first 50 headlines for controlled leaderboard comparison. (2) \textbf{Humor Transfer Bench (HTB)}: our new benchmark with 400 prompts (8 domains $\times$ 50 prompts) designed to evaluate a model's ability to generalize humor generation beyond standard headline-style inputs (see Section~\ref{htb_def}).

\subsection{Baselines}
We evaluate 15 total models: frontier systems (GPT-5, Kimi-K2, Gemini-2.5-Pro), open-weight general models (Qwen3-32B, GPT-OSS-120B, Base Qwen-7B), HumorGen variants (SFT, DPO, O-GRPO and their Think versions, plus HumorGen-Com-7B), and humor-specialized baselines (phi2-Humor~\cite{abbas2026phi2humor}, JokeGPT~\cite{jentzsch2023chatgpt}).

\subsection{Implementation Details}
All models were trained on NVIDIA H100 (80GB) GPUs using LoRA~\cite{hu2022lora} ($r$=16) with the Unsloth library. SFT ran for 3 epochs; DPO and O-GRPO for 5 epochs, both with early stopping (patience=2). Candidate ranking for the full pool consumed $\sim$132 H100 node-hours. For O-GRPO, groups of $G=24$ candidates per prompt maximize the advantage-weighted learning signal.

\subsection{Evaluation Protocols}

\paragraph{Ranking methodology:} 
For each pair of jokes (A, B), the LLM judge selects the funnier one (or tie), with presentation order randomized per match to mitigate position bias. We aggregate all outcomes into a contest matrix, fit a Bradley-Terry (BT) model~\cite{gao2025re,ajayi2026humorrank} via the MM algorithm to estimate latent ratings, and report Elo-scale ratings with 95\% bootstrap confidence intervals. To account for evaluation variance and ensure reproducibility, all bootstrap resampling operations are executed with a fixed random seed~\cite{ajayi2026humorrank}. For cross-judge Kendall comparisons, exact BT ties are broken deterministically by model-name order before rank correlation is computed; key model comparisons have non-overlapping confidence intervals and are statistically significant. We do not report reference-based metrics (e.g., BLEU, ROUGE) because humor generation is inherently open-ended and subjective, motivating pairwise preference-based evaluation.

\begin{enumerate}
    \item Automated Pairwise Evaluation: We report Llama-3.3-70B and Qwen 2.5 72B judged results for both benchmarks in the main paper: SemEval leaderboard (15 models, 5,250 pairwise comparisons) and HTB full benchmark (400 prompts, 15 models, 42,000 pairwise comparisons).
    \item Human Validation: Human evaluators judge 60 curated pairwise comparisons across 12 ablation categories; they are blinded to model identity and presentation order is randomized to mitigate position bias. We report inter-annotator agreement, LLM--human consensus, and correlation with automated BT ratings.
\end{enumerate}

\subsection{Humor Transfer Bench (HTB)}
\label{htb_def}
A key challenge in evaluating textual humor generation models is the absence of diverse benchmark datasets. The only existing benchmark for this task is the SemEval 2026 MWAHAHA test set, which uses news headlines as prompts, a single, narrow input distribution that limits assessment of model generalizability. To address this gap, we introduce the \textbf{Humor Transfer Bench (HTB)}, a new benchmark of 400 prompts spanning eight structurally distinct input domains (50 each): \textit{Neutral Facts}, \textit{Everyday Life}, \textit{Abstract Concepts}, \textit{Dialogic Quotations}, \textit{Scenario Inputs}, \textit{Analogical Prompts}, \textit{Direct Instructional}, and \textit{News Headlines}. No model is fine-tuned on HTB data; any performance gain on out-of-distribution domains therefore reflects genuine humor transfer. Full domain descriptions and design rationale are in Appendix~\ref{sec:appendix_htb_design}.

\section{Results}

\subsection{Model Performance}
\begin{figure}[t]
\centering
\includegraphics[width=\columnwidth]{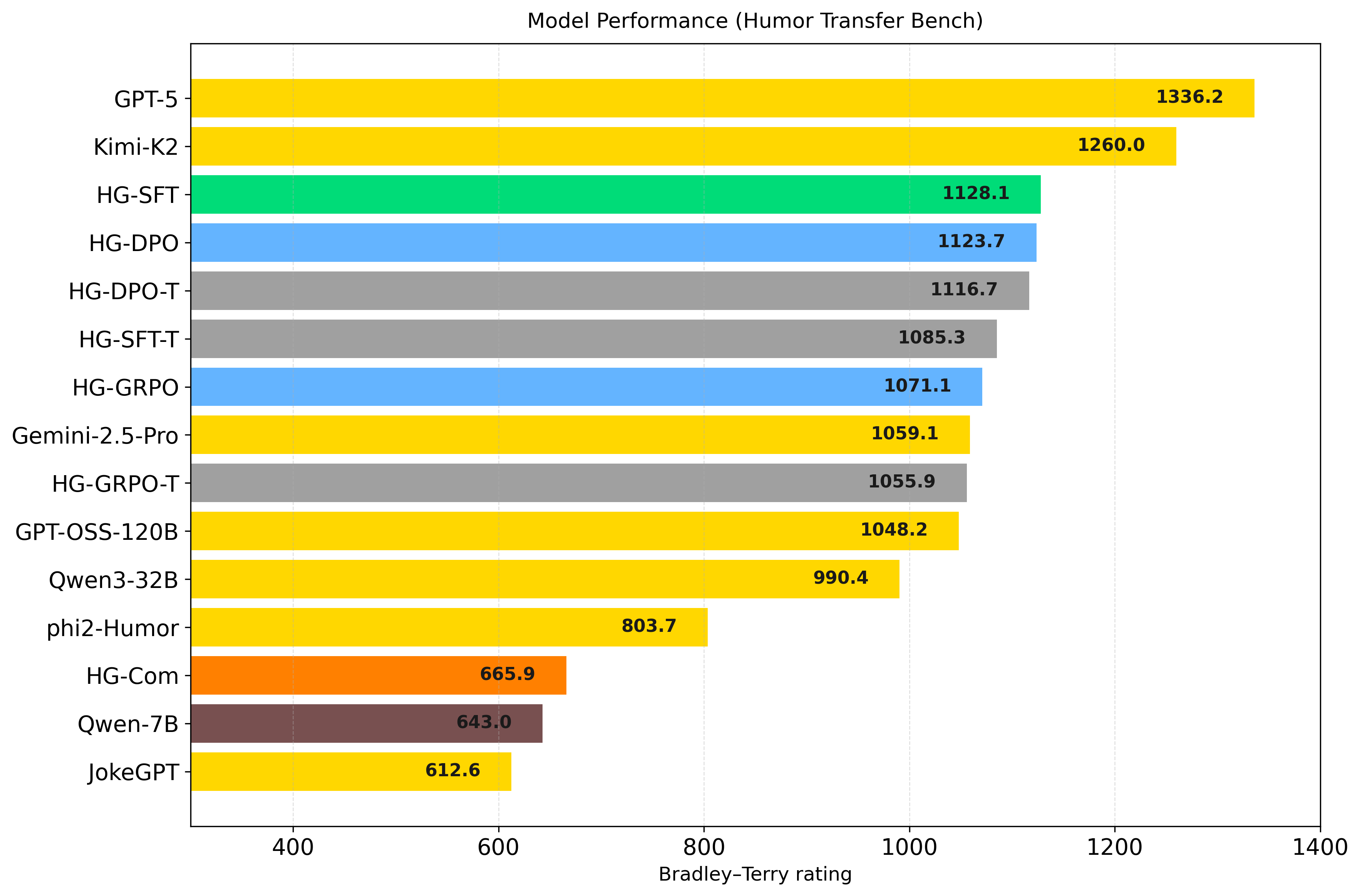}
\caption{HTB BT leaderboard (Llama judge, 400 prompts, 15 models). HumorGen-SFT and DPO-7B rank 3rd/4th, outperforming models 4--18$\times$ their size.}
\label{fig:htb_bt_main}
\end{figure}

\begin{figure}[t]
\centering
\includegraphics[width=\columnwidth]{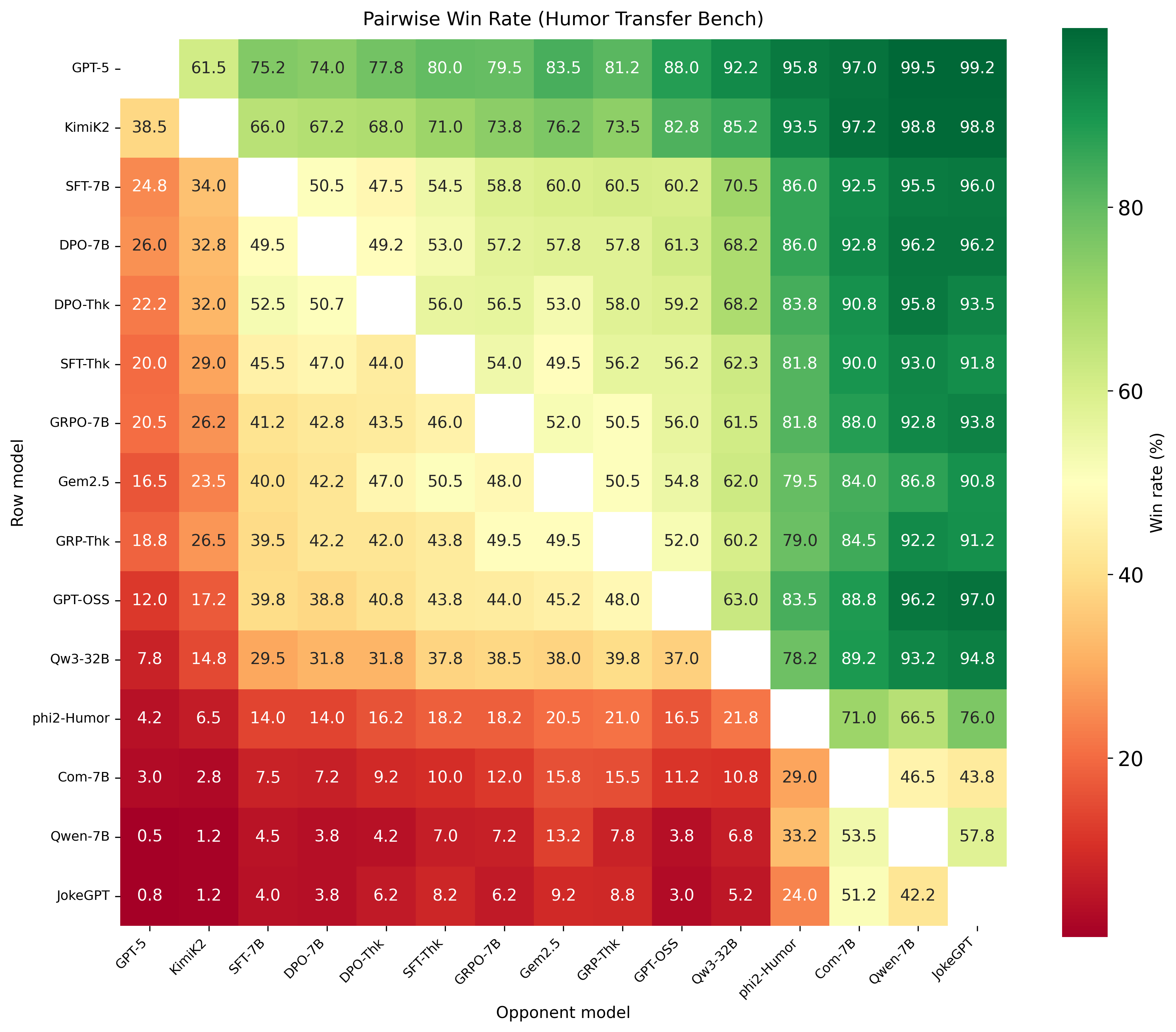}
\caption{HTB pairwise win-rate heatmap (Llama judge). Each cell shows the row model's win rate against the column model.}
\label{fig:htb_heatmap_main}
\end{figure}

We evaluate all 15 models on the Humor Transfer Bench under the Llama 3.3-70B judge (42,000 pairwise comparisons). Figure~\ref{fig:htb_bt_main} shows the Bradley-Terry leaderboard; Figure~\ref{fig:htb_heatmap_main} shows the head-to-head win-rate heatmap across all model pairs. \textbf{HumorGen-SFT-7B} ranks 3rd (BT~1128.14) and \textbf{HumorGen-DPO-7B} ranks 4th (BT~1123.72), ahead of Gemini-2.5-Pro (BT~1059.07), GPT-OSS-120B (BT~1048.19), and Qwen3-32B (BT~990.44), all of which are 4--18$\times$ larger. Humor-finetuned baselines phi2-Humor and JokeGPT rank 12th and 15th, confirming that domain specialization alone, without cognitive-driven data curation, does not generalize. To verify rating stability across evaluators, we run Qwen 2.5-72B as a second independent judge; the cross-judge Kendall $\tau = 0.8667$ ($p=1.54\times10^{-7}$) confirms that model ordering is robust across judges. We additionally evaluate both judges on the SemEval news-headline test set (15 models, 50 headlines), where cross-judge agreement reaches Kendall $\tau = 0.8286$ ($p=9.95\times10^{-7}$). Full leaderboard tables with 95\% CI and all Qwen-judge results are reported in Appendix~\ref{sec:appendix_leaderboards}. Overall, HumorGen 7B variants consistently outperform significantly larger models across unseen prompt domains, validating the generalization capacity of the Cognitive Synergy Framework.

\subsection{Preference Alignment}

We investigate whether preference alignment (DPO, O-GRPO) improves over SFT. Across both benchmarks and judges, SFT and DPO perform similarly, with negligible and inconsistent differences: on HTB (Llama judge), SFT (BT~1128.14) slightly outperforms DPO (BT~1123.72), while under the Qwen judge, DPO (BT~1138.05) narrowly exceeds SFT (BT~1132.41). In both cases, the 95\% confidence intervals overlap, indicating no statistically significant difference. O-GRPO consistently underperforms both SFT and DPO (e.g., BT~1071.13 on HTB under the Llama judge, rank 7th). This gap is attributable to a skewed advantage distribution in the O-GRPO training data (median advantage $= -0.479$), where a small number of high-quality samples dominate the group mean, causing most candidates to receive negative relative advantages. As a result, the optimization signal primarily penalizes weaker samples rather than reinforcing strong ones, leading to stable underperformance. A detailed analysis of this structural skew and our RankNorm ablation is provided in Appendix~\ref{sec:appendix_ogrpo}. These findings suggest a \textit{data quality ceiling}: when SFT data is already well-curated via CSF, additional preference optimization yields limited gains in this subjective humor generation setting.

\subsection{CSD and the Explainer Trap}
\label{sec:results_csd}
The ``explainer trap'' emerges when we train the 7B HumorGen variants to \textit{think} by applying Cognitive Synergy Distillation (CSD) so the student is trained on the teacher's \texttt{<think>} reasoning traces alongside the joke (Section~\ref{sec:csd}). On automatic leaderboards, Think variants \textit{usually} rank below their matched non-thinking counterparts (e.g., SFT-Think 6th vs.\ SFT 3rd on HTB under Llama; DPO-Think 5th vs.\ DPO 4th), but the effect is not uniform: DPO-Think ranks 3rd under the Qwen judge on HTB, ahead of DPO (4th). Distilling reasoning traces appears to bias outputs toward \textit{explaining} the joke rather than delivering it (Appendix~\ref{sec:appendix_think_vs_non_think}), consistent with the Qwen teacher model writing longer, more analytical jokes than the Kimi teacher model (Table~\ref{tab:cognitive_differences}). This extends prior work showing limited benefit of CoT for humor~\cite{zhong_lets_2024, tikhonov_humor_2024}: reasoning-augmented training does not reliably improve judged funniness in our setup.

\subsection{Comedian Adaptation}

To further probe the data quality thesis, we fine-tuned HumorGen-SFT on 998 stand-up jokes by comedian Shaun Eli~\cite{eli_expired_comedy}. Results show that performance regressed sharply (BT: 1083.9 $\to$ 653.1), as \textit{performance-native} stand-up is optimized for live delivery rather than written punchlines, confirming that data origin matters as much as data quantity. Full analysis of the comedian adaption experiment is provided in Appendix~\ref{sec:appendix_comedian_analysis}.

\subsection{Human Evaluation}
\label{sec:results_human_eval}

Three evaluators gave 180 blind pairwise judgments on 60 curated pairs (12 categories, 5 each) over 50 held-out headlines. Inter-annotator agreement among the humans was 31.7\% (one-third of pairs), reflecting humor's subjectivity. To assess the LLM judge's reliability against human preference, we additionally computed Krippendorff's alpha over the LLM judge together with the human evaluators, obtaining $\alpha = 0.412$ across three evaluators and $\alpha = 0.425$ across two, both in the moderate-agreement range and consistent with the highly subjective nature of humor evaluation. The LLM judge matched human consensus on 58.3\% of pairs (Gold) and individual votes at 52.4\% (Micro-Avg). In this ``Good vs.\ Good'' regime (high-quality outputs, no objectively worse option), 58.3\% indicates the judge captures shared preferences well above chance. More details of the human evaluation experiments are provided in Appendix~\ref{sec:appendix_human_eval}.

\section{Analysis}

\subsection{What Makes Jokes Win?}
For each HTB pairwise comparison, the Llama judge assigns structured tags to the winner (humor \textit{mechanisms} and \textit{delivery}) and to the loser (flaw tags). Table~\ref{tab:feature_contribution} aggregates all canonical tags across 41,976 non-tie comparisons from the full HTB Llama evaluation. Percentages denote the share of comparisons in which a tag was assigned (multiple tags per comparison are allowed).


\begin{table}[t]
\centering
\small
\begin{tabular}{@{}lr@{}}
\toprule
\textbf{Feature} & \textbf{\%} \\
\toprule
\multicolumn{2}{@{}l}{\textbf{Humor Mechanisms} \textit{(winner)}} \\
\midrule
\quad Incongruity        & 85.2 \\
\quad Absurdity          & 84.9 \\
\quad Surprise           & 57.2 \\
\quad Wordplay           & 42.0 \\
\quad Irony              & 22.5 \\
\quad Sarcasm            & 17.3 \\
\quad Narrative          & 16.5 \\
\quad Observational      & 14.7 \\
\toprule
\multicolumn{2}{@{}l}{\textbf{Delivery Features} \textit{(winner)}} \\
\midrule
\quad Punchline positioning & 96.1 \\
\quad Conciseness           & 72.0 \\
\quad Framing commitment    & 57.7 \\
\quad Escalation            & 33.0 \\
\quad Deadpan               & 11.7 \\
\quad Timing                &  4.9 \\
\toprule
\multicolumn{2}{@{}l}{\textbf{Loser Features} \textit{(loser)}} \\
\midrule
\quad Weak punchline    & 69.0 \\
\quad Clich\'e          & 64.9 \\
\quad Overexplained     & 22.3 \\
\quad Buried punchline  & 12.9 \\
\quad Confusing         &  8.8 \\
\quad Offensive         &  0.5 \\
\bottomrule
\end{tabular}
\caption{Judge-assigned tags on HTB (Llama; 41,976 non-tie comparisons). \% of comparisons tagged.}
\label{tab:feature_contribution}
\end{table}

Incongruity and absurdity dominate winning mechanisms, contributing to over 84\% of wins. Punchline positioning (96.1\%) and conciseness (72.0\%) dominate delivery features. Among losers, weak punchlines (69.0\%) and clich\'e (64.9\%) are the most frequent losing features. Overexplained (22.3\%) and buried punchlines (12.9\%) align with the explainer trap (Section~\ref{sec:results_csd}). Two recurring failure patterns emerge: (1)~\textit{generic punchlines} defaulting to safe, high-probability completions, and (2)~\textit{overextended setups} burying the joke. See Appendix~\ref{sec:appendix_examples} for qualitative examples.


\section{Conclusion}

We introduce the Cognitive Synergy Framework, which operationalizes psychological humor theories into six cognitive personas to generate diverse, high-quality humor data via Mixture-of-Thought. HumorGen model variants achieve strong performance among open-weight models and are competitive with frontier systems, outperforming significantly larger models, suggesting that targeted CSF data curation matters more than model scale for humor generation. Our central finding is a \textit{data quality ceiling}: when SFT data is diverse and well-curated, preference optimization (DPO, O-GRPO) yields no significant gains. We also observe that reasoning-augmented training can reduce judged funniness, while CSF-curated synthetic data outperforms human-written jokes when used in training. Human evaluation further supports that the LLM judge captures subtle preference differences even in subjective ``Good vs.\ Good'' pairwise comparisons among high-quality jokes. Overall, our results indicate that in modeling humor generation, performance gains are driven less by optimization complexity and more by the quality and structure of the underlying training data.


\section{Limitations}

This study defines a focused empirical scope that future work should broaden. Our analysis evaluates English, text-only humor on two benchmarks (SemEval-MWAHAHA and Humor Transfer Bench), motivating future evaluation across additional languages, domains, and multimodal settings. Human evaluation remains challenging due to the inherent subjectivity of humor and the existence of multiple valid outputs per prompt; accordingly, we employed a small annotator pool of three native English speakers evaluating 60 sampled pairwise joke comparisons. Larger and more demographically diverse evaluations could provide greater statistical power and deeper insights into variations in humor perception. Model comparisons include 15 selected models, comprising open-weight models ranging from 7B to 120B parameters across the Qwen, Llama, and GPT families, as well as specialized humor models. Due to computational constraints, our evaluation includes only a limited number of closed-weight models; future work could extend this analysis to a broader range of architectures and model scales. Finally, because humor interpretation is subjective and culturally dependent, generated jokes may be perceived as offensive or culturally insensitive by some audiences (see Appendix~\ref{sec:appendix_africa_jokes}), highlighting the need for further investigation of safety and cultural alignment in humor generation systems.

\section{Ethics Statement}
Humor generation risks producing offensive content. Our framework encourages creative mechanisms (e.g., wordplay, absurdity) over denigration or prejudice. Training data derives from the SEMEVAL 2026 MWAHAHA headlines and our synthetic candidates generated from those prompts; evaluation additionally uses HTB prompts for transfer assessment. Human evaluators were volunteers recruited by invitation; no payment was provided. All content in this work was written and verified by the authors. We used an LLM-based tool solely for grammar correction, language polishing, and LaTeX formatting assistance. The authors take full responsibility for the final content.

\clearpage
\section*{Acknowledgments}
This publication was developed as part of the Center for Inclusive Digital Transformation of Africa (CIDTA), and the Afretec Network which is managed by Carnegie Mellon University Africa and receives financial support from the Mastercard Foundation. The views expressed in this document are solely those of the authors and do not necessarily reflect those of Carnegie Mellon University or the Mastercard Foundation.


\bibliography{custom}

\clearpage
\appendix

\begin{center}
\large\bfseries
HumorGen: Cognitive Synergy for Humor Generation in Large Language Models via Persona-Based Distillation
\end{center}
\vspace{1.5em}

\section*{Appendix - Table of Contents}
\addcontentsline{toc}{section}{Contents of the Appendix}
\noindent
\textbf{\ref{sec:appendix_initial_models} Teacher Model Selection and Analysis} \dotfill \pageref{sec:appendix_initial_models}\\[0.5ex]
\textbf{\ref{sec:appendix_judge_selection} Pairwise LLM Judge Selection} \dotfill \pageref{sec:appendix_judge_selection}\\[0.5ex]
\textbf{\ref{sec:appendix_leaderboards} Benchmark Leaderboards} \dotfill \pageref{sec:appendix_leaderboards}\\
\quad\ref{sec:appendix_bootstrap} Bootstrap and CI Estimation Details \dotfill \pageref{sec:appendix_bootstrap}\\
\quad\ref{sec:appendix_htb_llama} HTB Full Results (Llama 3.3-70B Judge) \dotfill \pageref{sec:appendix_htb_llama}\\
\quad\ref{sec:appendix_htb_qwen} HTB Full Results (Qwen 2.5-72B Judge) \dotfill \pageref{sec:appendix_htb_qwen}\\
\quad\ref{sec:appendix_semeval_llama} SemEval Full Results (Llama 3.3-70B Judge) \dotfill \pageref{sec:appendix_semeval_llama}\\
\quad\ref{sec:appendix_semeval_qwen} SemEval Full Results (Qwen 2.5-72B Judge) \dotfill \pageref{sec:appendix_semeval_qwen}\\[0.5ex]
\textbf{\ref{sec:appendix_htb_design} HTB Design and Domain Descriptions} \dotfill \pageref{sec:appendix_htb_design}\\
\quad\ref{sec:appendix_htb_datacard} Humor Transfer Bench Data Card \dotfill \pageref{sec:appendix_htb_datacard}\\[0.5ex]
\textbf{\ref{sec:appendix_persona} Per-Persona Analysis} \dotfill \pageref{sec:appendix_persona}\\
\textbf{\ref{sec:appendix_training_distribution} Training Dataset Persona Distribution} \dotfill \pageref{sec:appendix_training_distribution}\\[0.5ex]
\textbf{\ref{sec:appendix_humor_theories} Psychological Theories of Humor} \dotfill \pageref{sec:appendix_humor_theories}\\[0.5ex]
\textbf{\ref{sec:appendix_hyperparameters} Training Details and Hyperparameters} \dotfill \pageref{sec:appendix_hyperparameters}\\
\quad\ref{sec:appendix_ogrpo} O-GRPO Objective \dotfill \pageref{sec:appendix_ogrpo}\\
\quad\quad\ref{sec:appendix_ranknorm} Ablation Study: Rank-Normalized Advantages \dotfill \pageref{sec:appendix_ranknorm}\\
\quad\ref{sec:appendix_hyperparams_config} Hyperparameter Configurations \dotfill \pageref{sec:appendix_hyperparams_config}\\
\quad\ref{sec:appendix_training_dynamics} Training Dynamics and Results \dotfill \pageref{sec:appendix_training_dynamics}\\
\quad\ref{sec:appendix_eval_loss} Evaluation Loss Trajectories \dotfill \pageref{sec:appendix_eval_loss}\\
\quad\ref{sec:appendix_comedian_hyperparams} Comedian Adaptation Hyperparameters \dotfill \pageref{sec:appendix_comedian_hyperparams}\\[0.5ex]
\textbf{\ref{sec:appendix_prompts} Full Persona Prompts} \dotfill \pageref{sec:appendix_prompts}\\[0.5ex]
\textbf{\ref{sec:appendix_persona_comparison} Immersive Persona Comparison} \dotfill \pageref{sec:appendix_persona_comparison}\\[0.5ex]
\textbf{\ref{sec:appendix_think_vs_non_think} Think vs.\ Non-Think} \dotfill \pageref{sec:appendix_think_vs_non_think}\\[0.5ex]
\textbf{\ref{sec:appendix_human_eval} Human Evaluation Details} \dotfill \pageref{sec:appendix_human_eval}\\[0.5ex]
\textbf{\ref{sec:appendix_evaluation_ui} Evaluation UI} \dotfill \pageref{sec:appendix_evaluation_ui}\\[0.5ex]
\textbf{\ref{sec:appendix_ranking_granularity} Humor Ranking Granularity} \dotfill \pageref{sec:appendix_ranking_granularity}\\[0.5ex]
\textbf{\ref{sec:appendix_comedian_analysis} Comedian Adaptation Analysis} \dotfill \pageref{sec:appendix_comedian_analysis}\\[0.5ex]
\textbf{\ref{sec:appendix_africa_jokes} Qualitative Output Examples: Out-of-Distribution Headlines} \dotfill \pageref{sec:appendix_africa_jokes}\\[0.5ex]
\textbf{\ref{sec:appendix_examples} Failure Mode Examples} \dotfill \pageref{sec:appendix_examples}
\vspace{1.5em}

\clearpage

\section{Teacher Model Selection and Analysis}
\label{sec:appendix_initial_models}

To construct a high-quality candidate pool for humor distillation, we initially experimented with several top-tier open-source models to serve as the teacher model. Our goal was to select a model capable of generating not just structurally coherent text, but subjectively funny jokes. Table~\ref{tab:teacher_models} summarizes our qualitative observations during this model selection phase. 

\begin{table}[!h]
\centering
\small
\resizebox{\columnwidth}{!}{%
\begin{tabular}{p{0.2\linewidth} p{0.35\linewidth} p{0.35\linewidth}}
\toprule
\textbf{Model} & \textbf{Rationale for Evaluation} & \textbf{Qualitative Observation} \\
\midrule
DeepSeek-67B & Top-tier open-source model available during initial development. & Generated jokes with poor comedic timing; subjectively less funny relative to Qwen-2.5-32B. \\
\addlinespace
Qwen-2.5-32B & Selected as a robust baseline due to its strong performance-to-size ratio and high instruction-following capabilities, which proved highly effective for structurally complex joke generation. & Generated structurally sound and elaborate jokes, though often lacked concise punchiness. \\
\addlinespace
Kimi-K2 & Explored specifically to introduce stylistic diversity alongside Qwen. & Generated consistently punchy, concise jokes with excellent comedic timing. \\
\bottomrule
\end{tabular}}
\caption{Observations during the teacher model selection phase.}
\label{tab:teacher_models}
\end{table}

Based on these observations, we recognized that while Qwen-2.5-32B generated high-quality setups, its outputs were consistently long. To introduce stylistic diversity and capture distinct cognitive strengths, we utilized an ensemble of both Qwen-2.5-32B and Kimi-K2. Analysis of their respective outputs confirmed that these architectural distinctions manifest directly in their generated styles (Table~\ref{tab:cognitive_differences}).

\begin{table}[!h]
\centering
\small
\resizebox{\columnwidth}{!}{%
\begin{tabular}{lll}
\toprule
\textbf{Feature} & \textbf{Qwen-2.5-32B} & \textbf{Kimi-K2} \\
\midrule
Average Word Count & $\sim$54.7 words & $\sim$37.5 words \\
Structural Style & Elaborate setups, detailed narratives & Concise setups, rapid punchlines \\
Cognitive Tone & Analytical and descriptive & Punchy, sharp, and conversational \\
\bottomrule
\end{tabular}}
\caption{Stylistic and cognitive differences between the two teacher models.}
\label{tab:cognitive_differences}
\end{table}

Consequently, both models contributed significantly to the final datasets. As shown in Table~\ref{tab:training_distribution}, the data distribution across all alignment corpora is highly balanced. This confirms that the ensemble effectively provided the student model with the best qualities of both teachers, ensuring robust stylistic variance across the generation pool.

\begin{table}[!h]
\centering
\small
\resizebox{\columnwidth}{!}{%
\begin{tabular}{lcc}
\toprule
\textbf{Dataset Partition} & \textbf{Qwen-2.5-32B Contribution} & \textbf{Kimi-K2 Contribution} \\
\midrule
SFT Training Corpus & 44.8\% & 55.2\% \\
DPO Chosen Pool & 40.9\% & 59.1\% \\
DPO Rejected Pool & 59.6\% & 40.4\% \\
O-GRPO Full Pool & 50.1\% & 49.9\% \\
\bottomrule
\end{tabular}}
\caption{Training data distribution contribution between the two teacher models across all alignment phases.}
\label{tab:training_distribution}
\end{table}

To examine whether this balanced contribution reflects comparable candidate quality rather than raw generation volume alone, we additionally aggregate the within-prompt rankings by teacher. Table~\ref{tab:teacher_post_ranking} shows that Kimi-K2 has a modest overall advantage, but Qwen-2.5-32B still produces the top-ranked joke for 45.2\% of prompts. Thus, neither teacher dominates the curated pool: Kimi contributes more concise, highly ranked candidates, while Qwen remains a substantial source of both preferred examples and longer-form stylistic variation.

\begin{table}[!h]
\centering
\small
\resizebox{\columnwidth}{!}{%
\begin{tabular}{lcc}
\toprule
\textbf{Post-ranking statistic} & \textbf{Kimi-K2} & \textbf{Qwen-2.5-32B} \\
\midrule
Mean Elo & 1003.9 & 996.2 \\
Mean finish rank ($\downarrow$) & 11.5 & 13.4 \\
Rank-1 wins & 603 (54.8\%) & 497 (45.2\%) \\
Share of top-3 positions & 63.6\% & 36.4\% \\
Share of bottom-half positions (\#11--24) & 46.2\% & 53.8\% \\
\bottomrule
\end{tabular}}
\caption{Post-ranking candidate quality by teacher across 1,100 prompts. Rankings are computed within each 24-candidate prompt group before constructing the SFT and preference-alignment corpora.}
\label{tab:teacher_post_ranking}
\end{table}

\section{Pairwise LLM Judge Selection}
\label{sec:appendix_judge_selection}

Before ranking the full candidate pool, we piloted several LLMs as pairwise humor judges. We assessed whether each model produced stable preferences, remained sensitive to comedic quality rather than surface characteristics such as response length, and avoided systematically favoring outputs from its own model family. Table~\ref{tab:curation_judge_selection} summarizes our qualitative observations. Llama-3.3-70B-Instruct was selected as the primary judge for training-data curation, while Qwen-2.5-72B was retained as an independent secondary judge for validating the resulting model rankings.

\begin{table}[!h]
\centering
\small
\resizebox{\columnwidth}{!}{%
\begin{tabular}{p{0.22\linewidth} p{0.58\linewidth} c}
\toprule
\textbf{Judge model} & \textbf{Observation} & \textbf{Selected} \\
\midrule
Llama-3.3-70B-Instruct &
Produced the most stable pairwise preferences and showed the strongest correspondence with human judgments in our blind calibration study. &
Yes (primary) \\
\addlinespace
Qwen-2.5-72B &
Produced rankings broadly consistent with the primary judge, while providing an independent model-family perspective for cross-judge validation. &
Yes (secondary) \\
\addlinespace
Qwen3-32B &
Produced less consistent ordering than the larger Qwen-2.5-72B judge in our preliminary comparisons. &
No \\
\addlinespace
GPT-OSS-120B &
Produced more variable preferences and was less consistent with the primary and secondary judges. &
No \\
\addlinespace
Gemini-2.5-Pro &
Provided generally coherent judgments, but its ordering was less consistent with the selected judges; its closed access also limited reproducibility. &
No \\
\bottomrule
\end{tabular}}
\caption{Qualitative comparison of candidate pairwise judges. Llama-3.3-70B-Instruct was selected as the primary curation judge, and Qwen-2.5-72B was selected as the secondary validation judge.}
\label{tab:curation_judge_selection}
\end{table}

We therefore used Llama-3.3-70B-Instruct to rank the candidate pool used to construct the SFT, DPO, and O-GRPO datasets. Qwen-2.5-72B was not used to curate training data; instead, it served as a secondary judge for testing whether the principal evaluation conclusions remained stable under an independent evaluator.

\setcounter{topnumber}{3}
\setcounter{bottomnumber}{3}
\setcounter{totalnumber}{6}
\renewcommand{\topfraction}{0.92}
\renewcommand{\bottomfraction}{0.85}
\renewcommand{\textfraction}{0.05}
\renewcommand{\floatpagefraction}{0.88}
\setcounter{dbltopnumber}{3}
\renewcommand{\dbltopfraction}{0.92}
\renewcommand{\dblfloatpagefraction}{0.88}
\tcbset{breakable, width=\columnwidth}
\newcommand{\appendixfigcaption}[2]{%
  \par\vspace{0.35em}%
  \captionof{figure}{#1}%
  \label{#2}%
  \vspace{0.45em}%
}

\section{Benchmark Leaderboards}
\label{sec:appendix_leaderboards}
Full leaderboard tables (BT ratings with 95\% bootstrap confidence intervals) and plots for all four evaluation conditions: HTB and SemEval, under both the Llama 3.3-70B and Qwen 2.5-72B judges.

\subsection{Bootstrap and CI Estimation Details}
\label{sec:appendix_bootstrap}
For each evaluation setting (HTB/SemEval $\times$ Llama/Qwen), we build a complete model--model outcome matrix from pairwise comparisons and fit Bradley--Terry ratings via the MM algorithm. We estimate uncertainty with non-parametric bootstrap resampling ($B=100$): each bootstrap replicate re-samples the evaluation units with replacement, reconstructs the full contest matrix, and re-fits the BT model. We report 95\% confidence intervals as percentile bounds (2.5th, 97.5th) over the bootstrap rating distribution for each model. This procedure is applied consistently to all leaderboard tables in this appendix.

\subsection{HTB Full Results (Llama 3.3-70B Judge)}
\label{sec:appendix_htb_llama}
\begin{figure}[!h]
\centering
\includegraphics[width=\columnwidth]{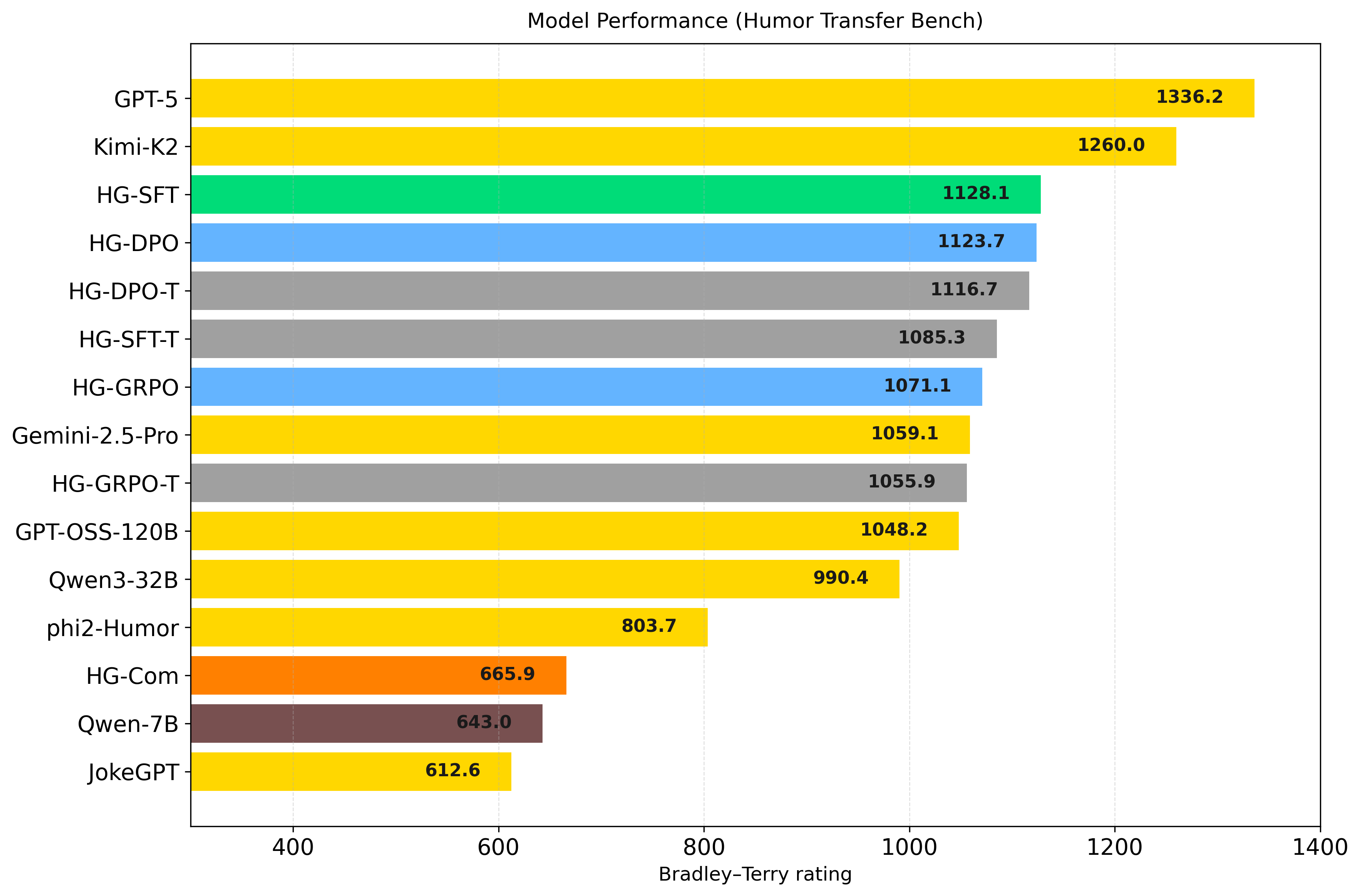}
\caption{HTB BT leaderboard, Llama judge (400 prompts, 15 models).}
\label{fig:htb_bt_appendix}
\end{figure}

\begin{figure}[!h]
\centering
\includegraphics[width=\columnwidth]{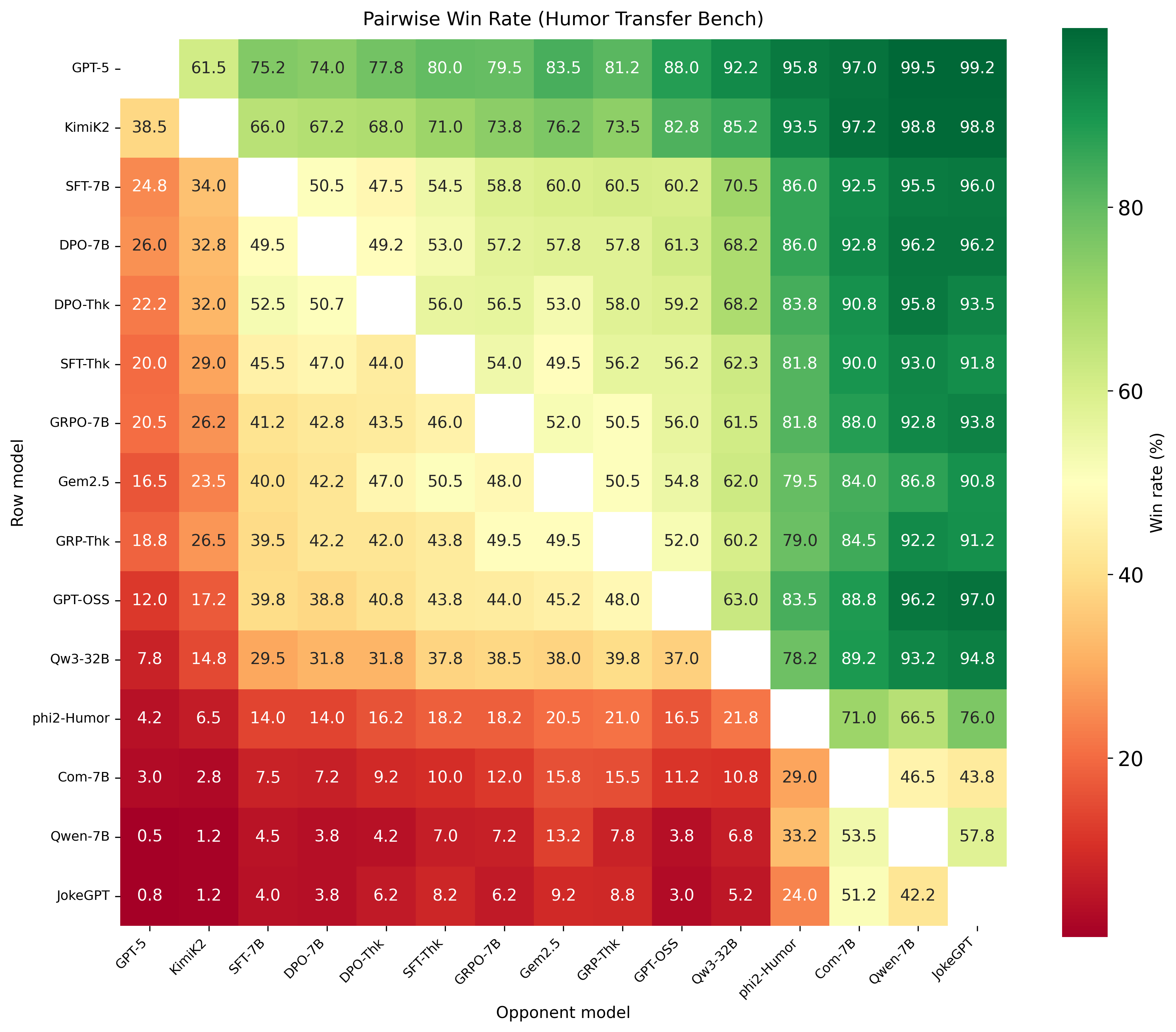}
\caption{HTB win-rate heatmap, Llama judge (row beats column \%).}
\label{fig:htb_winrate_heatmap_appendix}
\end{figure}

Table~\ref{tab:htb_llama_full} reports full BT ratings with 95\% bootstrap confidence intervals for all 15 models on HTB under the Llama 3.3-70B judge. Figure~\ref{fig:htb_bt_appendix} visualises the leaderboard; Figure~\ref{fig:htb_winrate_heatmap_appendix} shows the head-to-head win-rate matrix. HumorGen-SFT-7B and HumorGen-DPO-7B rank 3rd and 4th with non-overlapping CIs from all models below rank~6, confirming statistically significant separation from Gemini-2.5-Pro, GPT-OSS-120B, and all smaller baselines.

\begin{table}[!h]
\centering
\small
\resizebox{\columnwidth}{!}{%
\begin{tabular}{rlccc}
\toprule
\textbf{Rank} & \textbf{Model} & \textbf{BT Rating} & \textbf{95\% CI} & \textbf{Wins / Losses} \\
\midrule
1  & GPT-5                  & 1336.18 & [1323.3, 1348.3] & 4738 / 862  \\
2  & Kimi-K2                & 1259.98 & [1249.7, 1268.5] & 4362 / 1237 \\
3  & \textbf{HumorGen SFT-7B}   & \textbf{1128.14} & [1118.3, 1138.1] & 3565 / 2035 \\
4  & \textbf{HumorGen DPO-7B}   & \textbf{1123.72} & [1115.7, 1134.9] & 3536 / 2064 \\
5  & HumorGen DPO-Think-7B  & 1116.65 & [1107.9, 1127.1] & 3489 / 2110 \\
6  & HumorGen SFT-Think-7B  & 1085.31 & [1075.8, 1096.5] & 3281 / 2319 \\
7  & HumorGen GRPO-7B       & 1071.13 & [1060.8, 1080.1] & 3186 / 2414 \\
8  & Gemini-2.5-Pro         & 1059.07 & [1049.3, 1068.4] & 3104 / 2494 \\
9  & HumorGen GRPO-Think-7B & 1055.94 & [1043.8, 1066.8] & 3084 / 2516 \\
10 & GPT-OSS-120B           & 1048.19 & [1039.7, 1057.1] & 3032 / 2568 \\
11 & Qwen3-32B              &  990.44 & [981.4,  999.4]  & 2648 / 2952 \\
12 & phi2-Humor             &  803.72 & [794.5,  818.2]  & 1539 / 4060 \\
13 & HumorGen-Com-7B        &  665.93 & [645.5,  680.0]  &  897 / 4682 \\
14 & Base Qwen-7B           &  643.01 & [628.3,  658.0]  &  818 / 4781 \\
15 & JokeGPT                &  612.58 & [597.6,  627.4]  &  697 / 4882 \\
\bottomrule
\end{tabular}}
\caption{HTB leaderboard, Llama judge (400 prompts, 42,000 comparisons).}
\label{tab:htb_llama_full}
\end{table}

\subsection{HTB Full Results (Qwen 2.5-72B Judge)}
\label{sec:appendix_htb_qwen}
\begin{figure}[!h]
\centering
\includegraphics[width=\columnwidth]{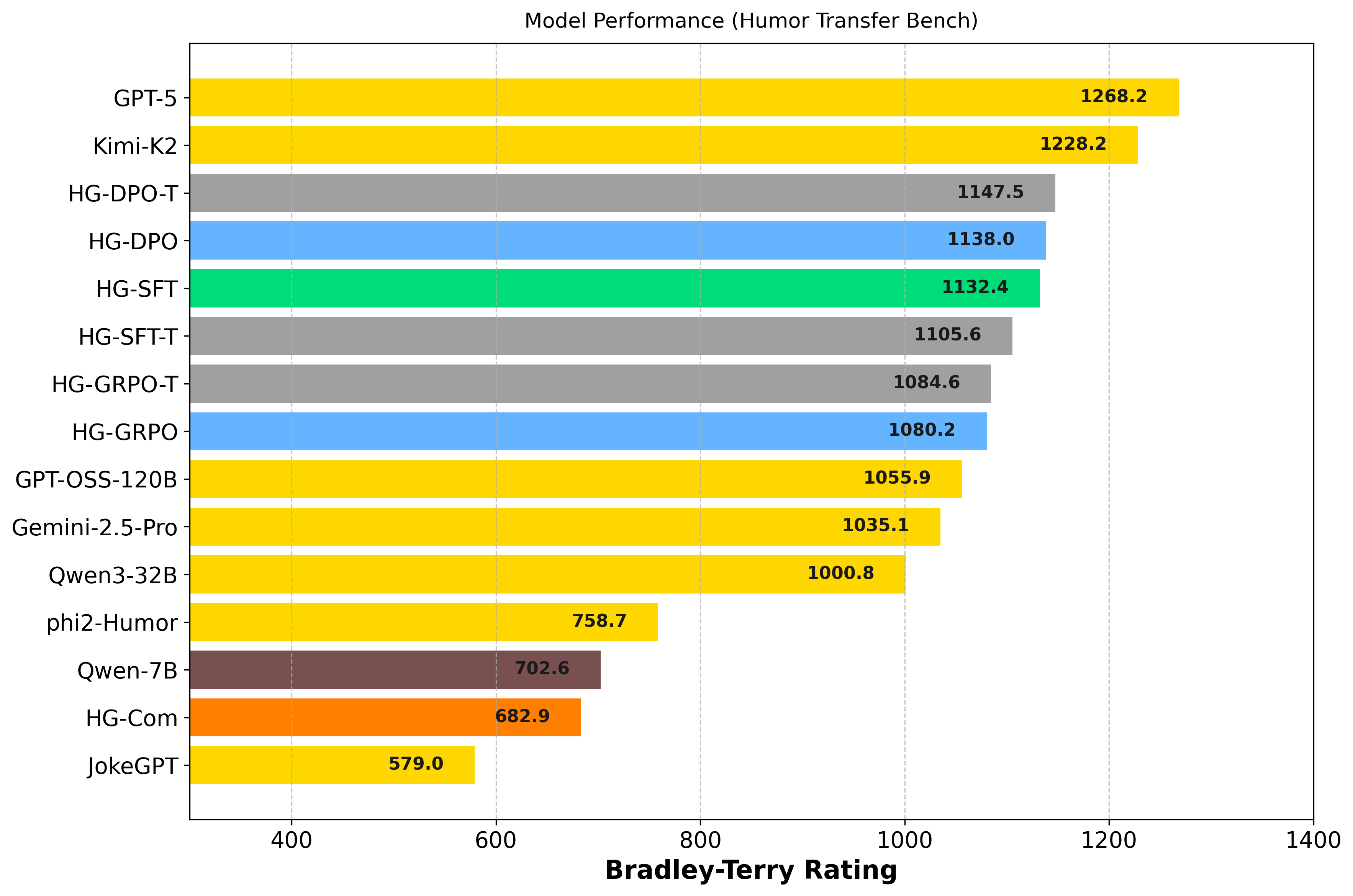}
\caption{HTB BT leaderboard, Qwen judge (400 prompts, 15 models).}
\label{fig:htb_qwen_bt_appendix}
\end{figure}

\begin{figure}[!h]
\centering
\includegraphics[width=\columnwidth]{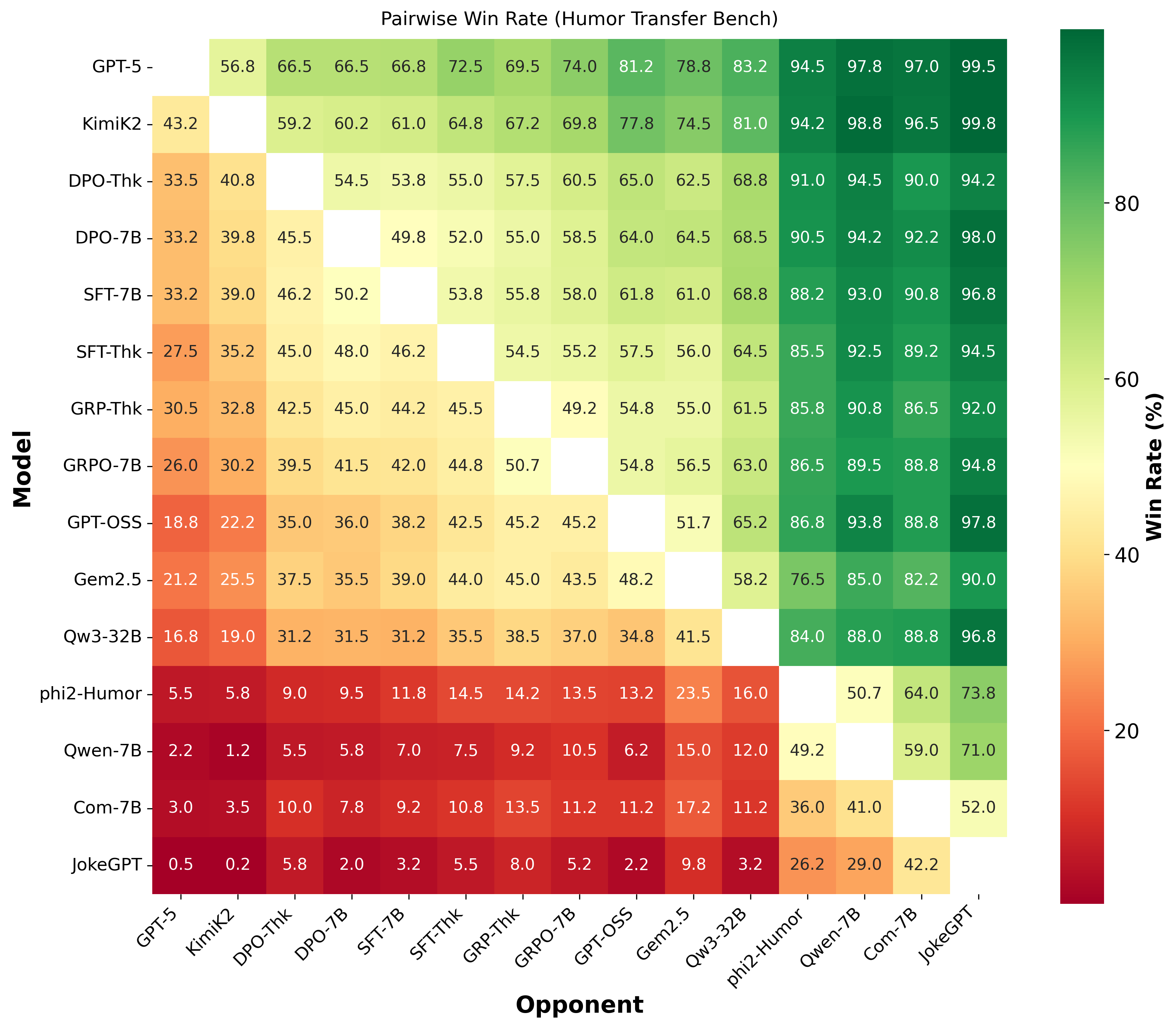}
\caption{HTB win-rate heatmap, Qwen judge (row beats column \%).}
\label{fig:htb_qwen_heatmap_appendix}
\end{figure}

Table~\ref{tab:htb_qwen_full} and Figures~\ref{fig:htb_qwen_bt_appendix}--\ref{fig:htb_qwen_heatmap_appendix} report HTB results under the Qwen 2.5-72B judge. The rank ordering is highly consistent with the Llama judge (Kendall $\tau = 0.8667$, $p=1.54\times10^{-7}$), confirming that the leaderboard is robust to judge choice. Notable difference: HumorGen-DPO-Think-7B rises to 3rd under Qwen, suggesting the Qwen judge is slightly more receptive to reasoning-augmented outputs.

\begin{table}[!h]
\centering
\small
\resizebox{\columnwidth}{!}{%
\begin{tabular}{rlccc}
\toprule
\textbf{Rank} & \textbf{Model} & \textbf{BT Rating} & \textbf{95\% CI} & \textbf{Wins / Losses} \\
\midrule
1  & GPT-5                  & 1268.25 & [1256.5, 1281.1] & 4418 / 1181 \\
2  & Kimi-K2                & 1228.24 & [1217.3, 1239.4] & 4192 / 1408 \\
3  & HumorGen DPO-Think-7B  & 1147.52 & [1136.8, 1158.8] & 3686 / 1914 \\
4  & \textbf{HumorGen DPO-7B}   & \textbf{1138.05} & [1130.3, 1147.2] & 3623 / 1976 \\
5  & \textbf{HumorGen SFT-7B}   & \textbf{1132.41} & [1121.0, 1141.2] & 3586 / 2014 \\
6  & HumorGen SFT-Think-7B  & 1105.57 & [1094.0, 1115.2] & 3406 / 2194 \\
7  & HumorGen GRPO-Think-7B & 1084.61 & [1075.2, 1097.0] & 3264 / 2336 \\
8  & HumorGen GRPO-7B       & 1080.19 & [1070.3, 1090.1] & 3234 / 2366 \\
9  & GPT-OSS-120B           & 1055.91 & [1046.4, 1065.5] & 3069 / 2531 \\
10 & Gemini-2.5-Pro         & 1035.10 & [1025.1, 1046.1] & 2926 / 2670 \\
11 & Qwen3-32B              & 1000.83 & [991.1,  1011.4] & 2698 / 2901 \\
12 & phi2-Humor             &  758.74 & [746.6,  769.0]  & 1300 / 4300 \\
13 & Base Qwen-7B           &  702.60 & [689.6,  715.1]  & 1046 / 4554 \\
14 & HumorGen-Com-7B        &  682.92 & [666.9,  692.9]  &  951 / 4624 \\
15 & JokeGPT                &  579.04 & [563.9,  596.0]  &  573 / 5003 \\
\bottomrule
\end{tabular}}
\caption{HTB leaderboard, Qwen judge (400 prompts, 42,000 comparisons). Kendall $\tau = 0.8667$ vs.\ Llama judge.}
\label{tab:htb_qwen_full}
\end{table}

\subsection{SemEval Full Results (Llama 3.3-70B Judge)}
\label{sec:appendix_semeval_llama}
\begin{figure}[!h]
\centering
\includegraphics[width=\columnwidth]{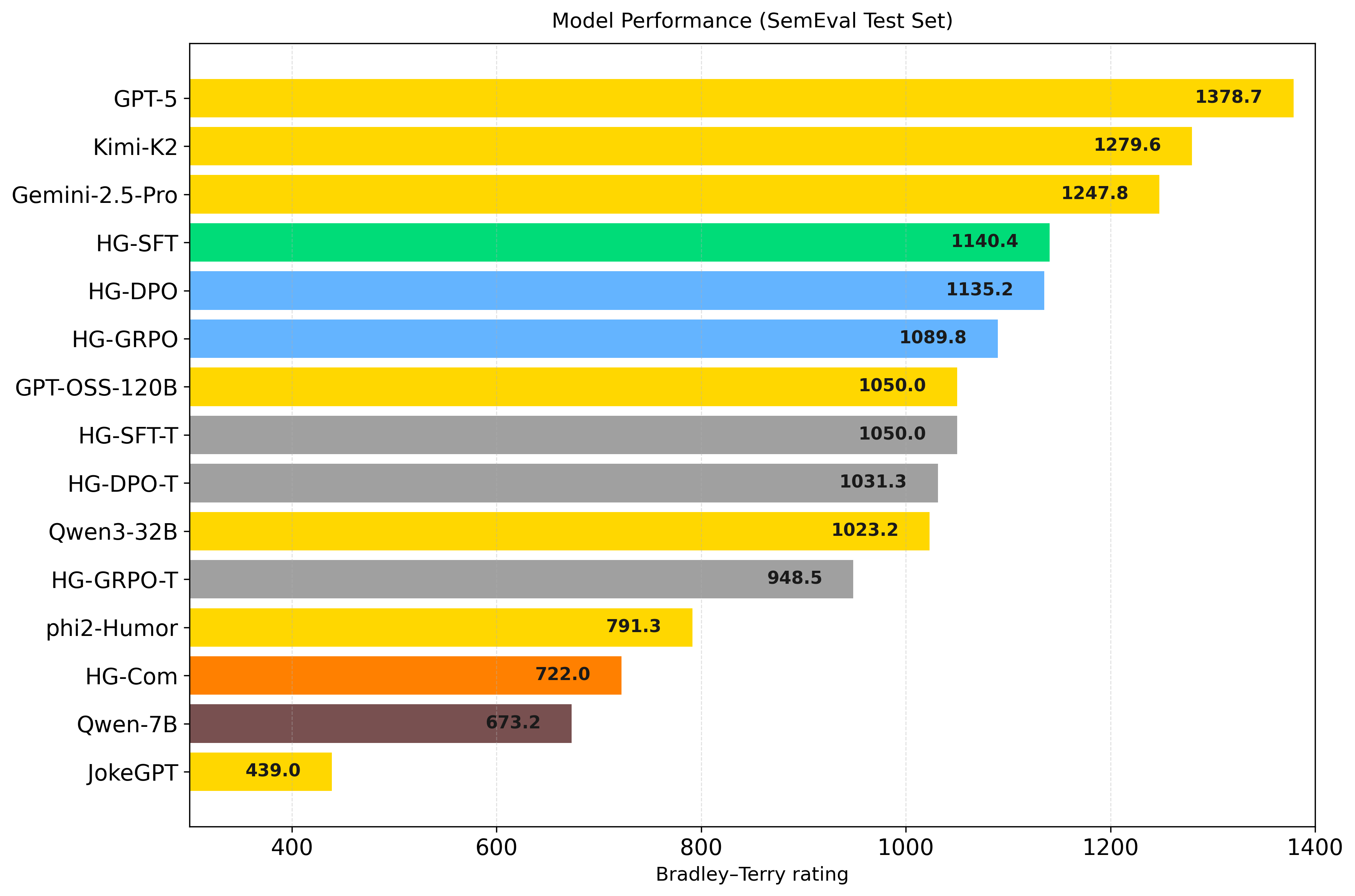}
\caption{SemEval BT leaderboard, Llama judge (50 headlines, 15 models).}
\label{fig:sem_llama_bt_appendix}
\end{figure}

\begin{figure}[!h]
\centering
\includegraphics[width=\columnwidth]{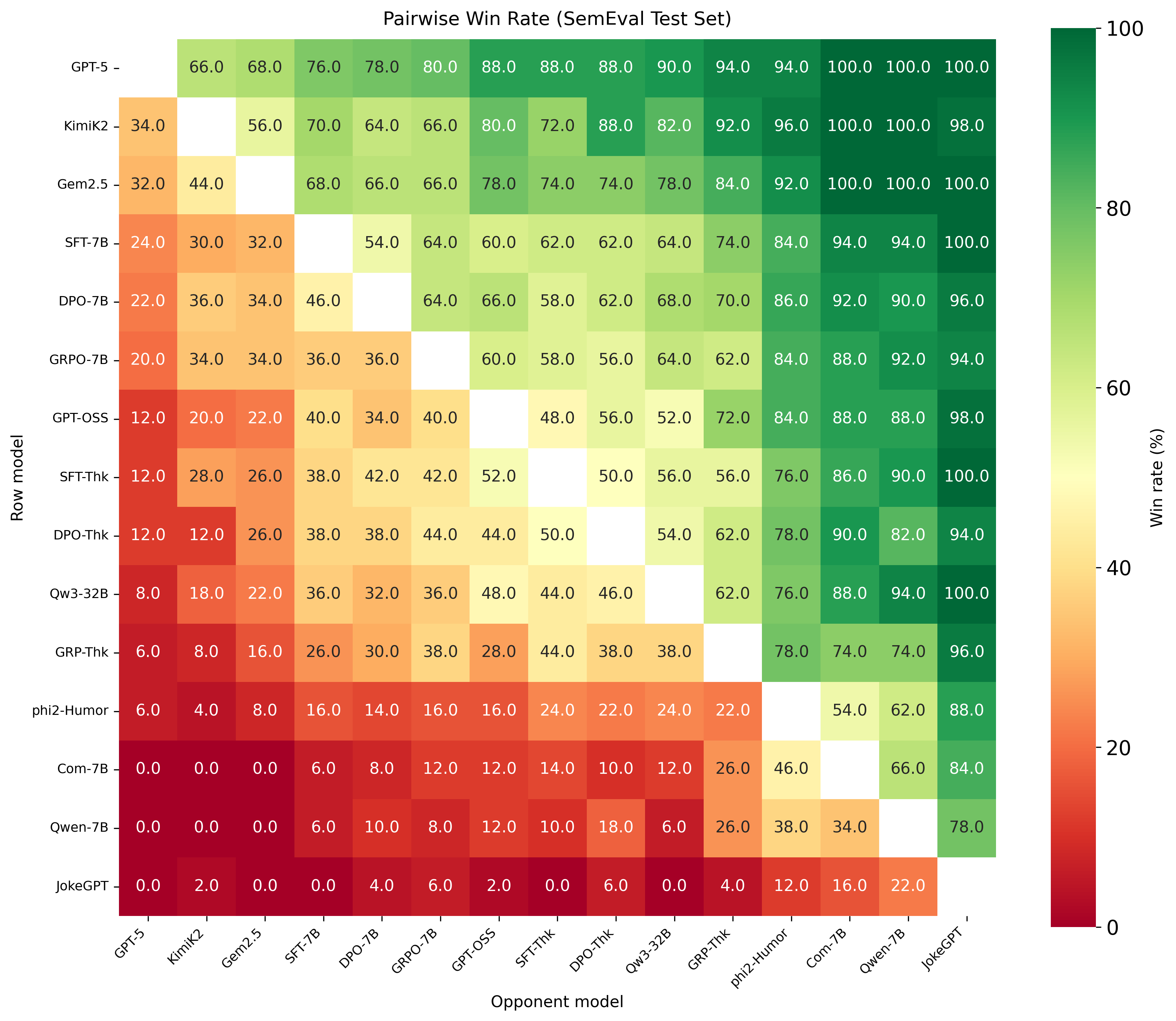}
\caption{SemEval win-rate heatmap, Llama judge (row beats column \%).}
\label{fig:sem_llama_heatmap_appendix}
\end{figure}

Table~\ref{tab:semeval_llama_full} and Figures~\ref{fig:sem_llama_bt_appendix}--\ref{fig:sem_llama_heatmap_appendix} report results on the SemEval 2026 MWAHAHA test set (first 50 headlines, 15 models, 5,250 pairwise comparisons) under the Llama 3.3-70B judge. HumorGen-SFT-7B and HumorGen-DPO-7B rank 4th and 5th, behind the three frontier systems (GPT-5, Kimi-K2, Gemini-2.5-Pro) and ahead of all open-weight general models. The gap between SFT and DPO (4.88 BT points) is within overlapping confidence intervals, confirming no statistically meaningful difference.

\begin{table}[!h]
\centering
\small
\resizebox{\columnwidth}{!}{%
\begin{tabular}{rlccc}
\toprule
\textbf{Rank} & \textbf{Model} & \textbf{BT Rating} & \textbf{95\% CI} & \textbf{Wins / Losses} \\
\midrule
1  & GPT-5                  & 1378.73 & [1346.1, 1421.5] & 605 / 94  \\
2  & Kimi-K2                & 1279.63 & [1245.0, 1322.1] & 549 / 151 \\
3  & Gemini-2.5-Pro         & 1247.80 & [1212.4, 1279.7] & 528 / 172 \\
4  & \textbf{HumorGen SFT-7B}   & \textbf{1140.37} & [1107.9, 1173.2] & 449 / 251 \\
5  & \textbf{HumorGen DPO-7B}   & \textbf{1135.25} & [1101.9, 1160.2] & 445 / 255 \\
6  & HumorGen GRPO-7B       & 1089.84 & [1060.7, 1114.1] & 409 / 291 \\
7  & GPT-OSS-120B           & 1049.99 & [1019.8, 1081.5] & 377 / 323 \\
8  & HumorGen SFT-Think-7B  & 1049.99 & [1016.3, 1084.5] & 377 / 323 \\
9  & HumorGen DPO-Think-7B  & 1031.30 & [1002.1, 1058.0] & 362 / 338 \\
10 & Qwen3-32B              & 1023.18 & [997.0,  1046.0] & 355 / 344 \\
11 & HumorGen GRPO-Think-7B &  948.51 & [914.7,  982.6]  & 297 / 403 \\
12 & phi2-Humor             &  791.32 & [751.8,  826.1]  & 188 / 512 \\
13 & HumorGen-Com-7B        &  721.97 & [682.3,  750.8]  & 148 / 552 \\
14 & Base Qwen-7B           &  673.16 & [643.1,  718.0]  & 123 / 577 \\
15 & JokeGPT                &  438.97 & [384.3,  500.2]  &  37 / 663 \\
\bottomrule
\end{tabular}}
\caption{SemEval leaderboard, Llama judge (50 headlines, 15 models, 5,250 comparisons).}
\label{tab:semeval_llama_full}
\end{table}

\subsection{SemEval Full Results (Qwen 2.5-72B Judge)}
\label{sec:appendix_semeval_qwen}
\begin{figure}[!h]
\centering
\includegraphics[width=\columnwidth]{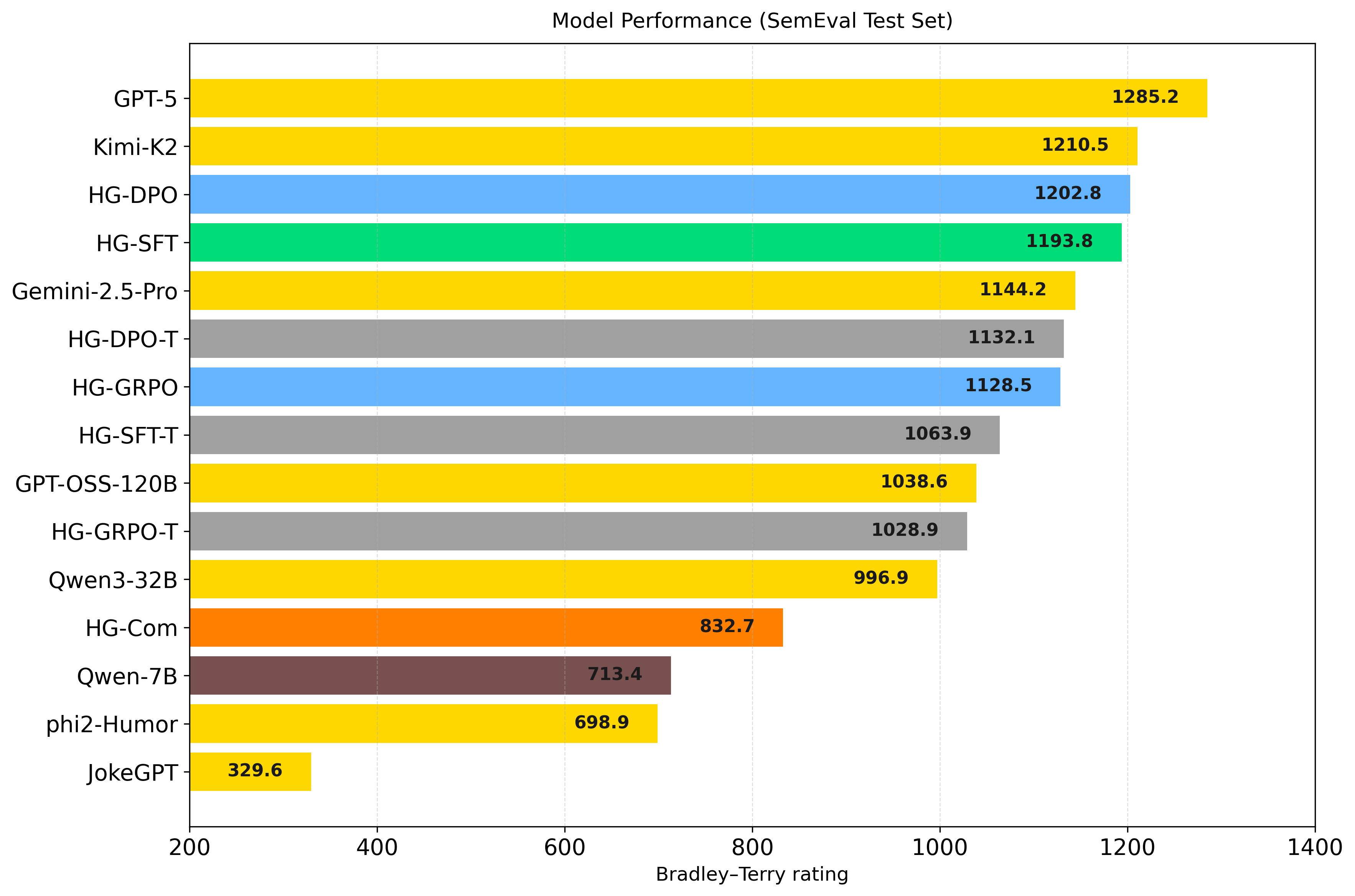}
\caption{SemEval BT leaderboard, Qwen judge (50 headlines, 15 models).}
\label{fig:sem_qwen_bt_appendix}
\end{figure}

\begin{figure}[!h]
\centering
\includegraphics[width=\columnwidth]{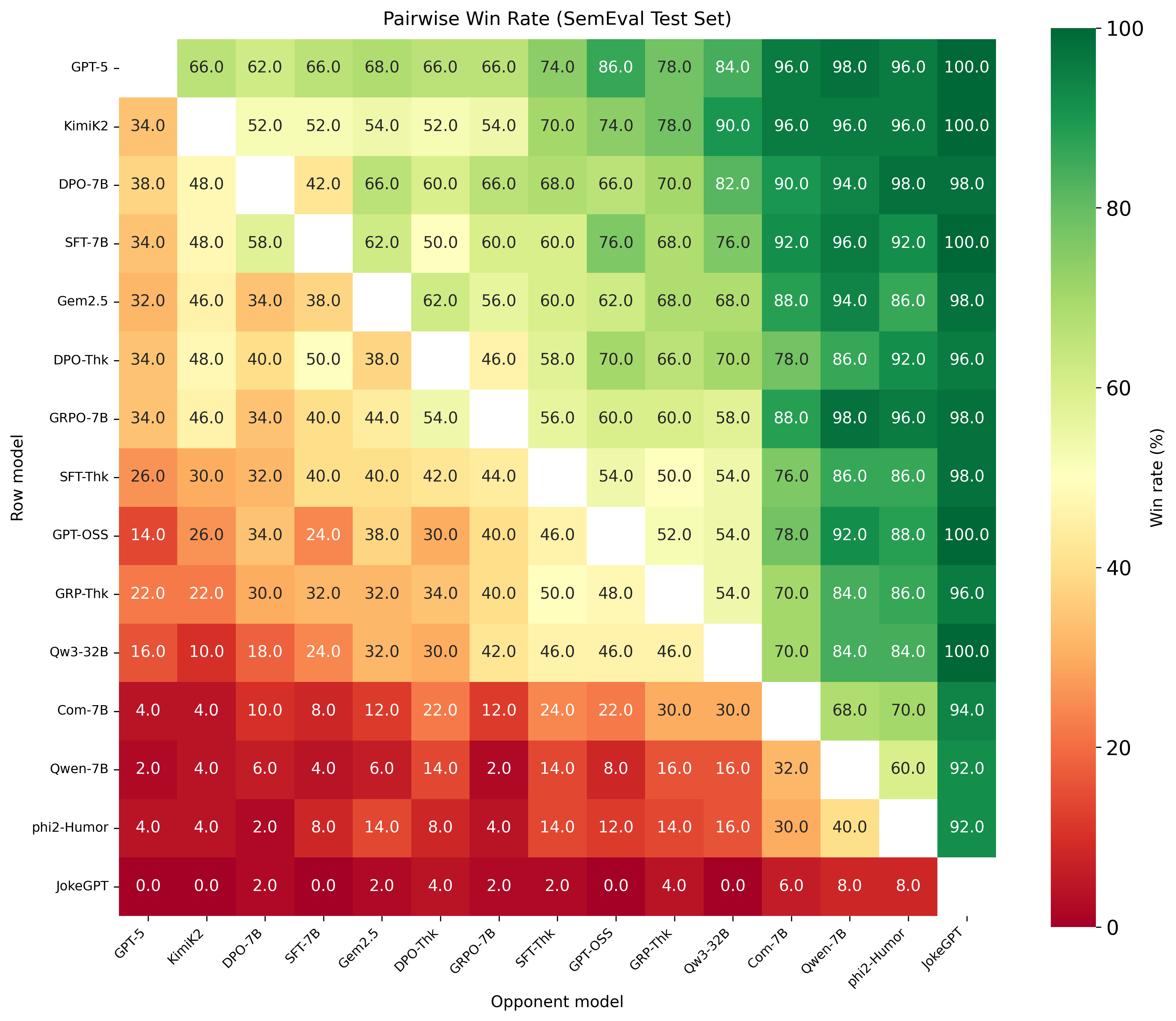}
\caption{SemEval win-rate heatmap, Qwen judge (row beats column \%).}
\label{fig:sem_qwen_heatmap_appendix}
\end{figure}

Table~\ref{tab:semeval_qwen_full} and Figures~\ref{fig:sem_qwen_bt_appendix}--\ref{fig:sem_qwen_heatmap_appendix} report SemEval results under the Qwen 2.5-72B judge. Rank ordering is consistent with the Llama judge at Kendall $\tau = 0.8286$ ($p=9.95\times10^{-7}$). Under Qwen, HumorGen-DPO-7B rises to 3rd (BT~1202.76) and HumorGen-SFT-7B to 4th (BT~1193.81), both ahead of Gemini-2.5-Pro, consistent with the HTB Qwen pattern of the Qwen judge rating DPO-aligned outputs slightly higher.

\begin{table}[!h]
\centering
\small
\resizebox{\columnwidth}{!}{%
\begin{tabular}{rlcc}
\toprule
\textbf{Rank} & \textbf{Model} & \textbf{BT Rating} & \textbf{95\% CI} \\
\midrule
1  & GPT-5                  & 1285.15 & [1258.2, 1320.8] \\
2  & Kimi-K2                & 1210.51 & [1188.5, 1233.9] \\
3  & \textbf{HumorGen DPO-7B}   & \textbf{1202.76} & [1175.5, 1227.0] \\
4  & \textbf{HumorGen SFT-7B}   & \textbf{1193.81} & [1162.0, 1228.2] \\
5  & Gemini-2.5-Pro         & 1144.23 & [1114.3, 1171.8] \\
6  & HumorGen DPO-Think-7B  & 1132.13 & [1103.0, 1170.3] \\
7  & HumorGen GRPO-7B       & 1128.51 & [1106.3, 1159.4] \\
8  & HumorGen SFT-Think-7B  & 1063.88 & [1034.9, 1091.8] \\
9  & GPT-OSS-120B           & 1038.60 & [1009.1, 1065.9] \\
10 & HumorGen GRPO-Think-7B & 1028.90 & [997.6,  1053.3] \\
11 & Qwen3-32B              &  996.92 & [968.0,  1030.5] \\
12 & HumorGen-Com-7B        &  832.69 & [802.3,  862.2]  \\
13 & Base Qwen-7B           &  713.39 & [674.7,  745.2]  \\
14 & phi2-Humor             &  698.89 & [657.3,  739.2]  \\
15 & JokeGPT                &  329.63 & [237.7,  385.6]  \\
\bottomrule
\end{tabular}}
\caption{SemEval leaderboard, Qwen judge (50 headlines, 15 models). Kendall $\tau = 0.8286$ vs.\ Llama judge.}
\label{tab:semeval_qwen_full}
\end{table}

\paragraph{Cross-benchmark observations.}
Three consistent patterns emerge across all four evaluation conditions (HTB and SemEval, Llama and Qwen judges). First, \textbf{HumorGen-SFT-7B and HumorGen-DPO-7B cluster tightly at ranks 3--5}, with BT differences between them within confidence interval overlap in every condition, consistent with the data quality ceiling finding in the main paper. Second, \textbf{Think variants most often trail their matched non-thinking counterparts} on automatic rankings (e.g., all three pairs on HTB--Llama and SemEval--Llama), though DPO-Think outranks DPO under the Qwen judge on HTB. Third, \textbf{humor-specialized baselines phi2-Humor and JokeGPT consistently rank 12th--15th} across both benchmarks and both judges, confirming that fine-tuning on humor data alone, without cognitive-driven curation, does not yield competitive performance. The high cross-judge agreement ($\tau \geq 0.83$ on both benchmarks) indicates these orderings are stable and not an artifact of judge choice.

\clearpage

\section{Humor Transfer Bench (HTB) Design and Domain Descriptions}
\label{sec:appendix_htb_design}

The Humor Transfer Bench (HTB) tests whether humor generation models generalize beyond the single prompt distribution present in existing benchmarks. HTB holds the task constant (\textit{Generate a funny joke given this prompt}) while varying the input domain. Each domain is defined by a distinct syntactic and pragmatic structure, making domain membership an observable property rather than a subjective label. The eight domains are described in Table~\ref{tab:htb_domains}.

\begin{table}[!h]
\centering
\small
\resizebox{\columnwidth}{!}{%
\begin{tabular}{@{}c l l p{5.5cm}@{}}
\toprule
\textbf{Dom.} & \textbf{Name} & \textbf{Status} & \textbf{Description and Example} \\
\midrule
A & Neutral Facts        & OOD              & Surprising declarative facts. \textit{``Bananas are technically berries.''} \\
B & Everyday Life        & OOD              & First-person observational statements. \textit{``My printer hates me.''} \\
C & Abstract Concepts    & OOD              & Bare noun-phrase prompts. \textit{``Bureaucratic nostalgia.''} \\
D & Dialogic Quotations  & OOD              & Attributed utterances. \textit{``The fridge said: I'm judging you.''} \\
E & Scenario Inputs      & OOD              & Situational irony setups. \textit{``A vampire is working in HR.''} \\
F & Analogical Prompts   & OOD              & Cross-domain relational mappings. \textit{``Life is a subscription service you never read the terms for.''} \\
G & Direct Instructional & Near-distrib.    & Directive format closest to training. \textit{``Make a joke about X.''} \\
H & News Headlines       & Reference News   & 50 real-world BBC headlines. \\
\bottomrule
\end{tabular}}
\caption{HTB domain descriptions used in evaluation.}
\label{tab:htb_domains}
\end{table}

\subsection{Humor Transfer Bench Data Card}
\label{sec:appendix_htb_datacard}
\begin{table}[!h]
\centering
\small
\begin{tabular}{@{}p{0.38\columnwidth}p{0.56\columnwidth}@{}}
\toprule
\textbf{Field} & \textbf{Value} \\
\midrule
Dataset name/version & Humor Transfer Bench (HTB), v1 \\
Primary objective & Measure humor transfer under prompt-domain shift while holding the generation task fixed \\
Task/instruction & Text-to-text humor generation (\textit{Generate a funny joke given this prompt}) \\
File format/encoding & TSV, UTF-8, Unix line endings \\
Schema (columns) & \texttt{id}, \texttt{domain}, \texttt{prompt} \\
Identifier format & \texttt{HTB\_\{DOMAIN\}\_\{NNN\}} (e.g., \texttt{HTB\_A\_001}) \\
Total size & 400 prompts (8 domains $\times$ 50 prompts) \\
Domain inventory & A: Neutral Facts (50); B: Everyday Life (50); C: Abstract Concepts (50); D: Dialogic Quotations (50); E: Scenario Inputs (50); F: Analogical Prompts (50); G: Direct Instructional Prompts (50); H: News Headlines (50) \\
Domain definition & Each category is defined by observable prompt structure; detailed examples are listed in Table~\ref{tab:htb_domains} \\
Data source & Domains A--G are HTB-authored prompts; Domain H uses 50 real-world BBC headlines \\
License/provenance & HTB-authored prompts are released under CC BY 4.0; Domain H uses publicly available news headline text \\
Exclusions & No canonical joke templates (e.g., knock-knock / walks-into-a-bar / chicken-road) in prompts \\
Training usage & HTB is evaluation-only in this paper (no model is fine-tuned on HTB prompts) \\
Evaluation usage & 15 models, pairwise judging under Llama-3.3-70B and Qwen-2.5-72B \\
\bottomrule
\end{tabular}
\caption{HTB data card aligned with the benchmark design specification.}
\label{tab:htb_data_card}
\end{table}

HTB is structured as eight equal-size prompt categories so that models can be compared consistently across varied humor-input styles.

\newpage
\clearpage
\section{Per-Persona Analysis}
\label{sec:appendix_persona}

A key design property of the Cognitive Synergy Framework is that \textit{no single persona is hand-picked as the ``best''}. All six personas generate candidates, the judge ranks them, and the Elo-based curation process decides which jokes enter training. Table~\ref{tab:persona_wins} reports three independent measures of curation-stage performance, all derived from the HumorRank pairwise tournament over 28,789 candidates across 1,200 prompts.

\begin{itemize}
  \item \textbf{Pairwise win rate:} proportion of individual 1v1 match-ups a persona's jokes won (source: full match history).
  \item \textbf{Mean rank (of 24):} average within-prompt rank across all 1,200 prompts (rank 1 = best in group; source: per-prompt rankings).
  \item \textbf{Top-3 rate:} proportion of prompts in which the persona produced a top-3 ranked joke.
\end{itemize}

All three metrics tell the same story and reinforce each other: \textbf{Neurotic} leads across every measure, while \textbf{Wordsmith} consistently trails. Crucially, all six personas remain competitive across different prompts. Wordsmith, despite ranking lowest, still achieves a top-3 placement on 3.1\% of prompts, and all personas contribute to the final training corpora (see Table~\ref{tab:training_persona_distribution}).

\begin{table}[H]
\centering
\small
\resizebox{\columnwidth}{!}{%
\begin{tabular}{@{}lrrr@{}}
\toprule
\textbf{Persona} & \textbf{Pairwise win rate} & \textbf{Mean rank (of 24)} & \textbf{Top-3 rate} \\
\midrule
Neurotic  & 63.4\% & 9.3  & 26.6\% \\
Absurdist & 55.8\% & 11.3 & 17.6\% \\
Cynic     & 55.2\% & 11.4 & 15.9\% \\
Observer  & 49.1\% & 12.5 &  6.2\% \\
Optimist  & 41.1\% & 14.4 &  5.7\% \\
Wordsmith & 34.9\% & 16.1 &  3.1\% \\
\bottomrule
\end{tabular}}
\caption{Per-persona curation-stage performance across three independent metrics. \textit{Pairwise win rate}: fraction of 1v1 judge comparisons won. \textit{Mean rank}: average within-prompt rank out of 24 candidates (lower = better). \textit{Top-3 rate}: fraction of prompts where the persona placed in the top 3. All three metrics converge on the same ordering.}
\label{tab:persona_wins}
\end{table}

The ordering reflects the humor mechanisms each persona draws on. Neurotic and Absurdist generate high incongruity and surprise through escalating anxiety and surreal logic respectively, consistent with the dominant winning tags in Table~\ref{tab:feature_contribution}. Cynic relies on satire and superiority framing, which scores similarly to Absurdist. Observer occupies the boundary just below 50\%, producing reliably solid jokes that rarely dominate. Optimist and Wordsmith trail because wholesome reframings lack edge, and forced wordplay is penalised by the judge. The consistency of the ordering across all three metrics confirms it is not an artefact of any single evaluation perspective.

\section{Training Dataset Persona Distribution}
\label{sec:appendix_training_distribution}

We trace each example in the SFT, DPO, and O-GRPO corpora back to its originating cognitive persona in the generation pool. Persona labels are taken from the raw CSF candidate pool and joined to the final alignment files by exact joke-text match (100\% coverage on the v4 corpora). Table~\ref{tab:training_persona_distribution} summarizes the resulting composition.

The raw candidate pool is balanced across all six personas ($\sim$16.7\% each). Elo-based selection for SFT (top 10 per prompt) and DPO (top 5 vs.\ bottom 5) shifts mass toward \textbf{Neurotic}, \textbf{Absurdist}, and \textbf{Cynic} on the preferred side, which directly reflects their higher mean ranks and top-3 rates in Table~\ref{tab:persona_wins}, while \textbf{Wordsmith} and \textbf{Optimist} are more frequent among rejected DPO examples. O-GRPO retains all 24 candidates per prompt, preserving uniform persona coverage before advantage weighting. Taken together, Tables~\ref{tab:persona_wins} and \ref{tab:training_persona_distribution} show that CSF does \textit{not} select a single dominant persona: the student model is trained on a distribution spanning all six cognitive styles, weighted by judge-assessed quality rather than by design.

\begin{table}[H]
\centering
\small
\caption{Persona composition of the alignment training corpora. Counts are followed by row percentages within each column (each column sums to 100\%). All six personas appear in every corpus.}
\label{tab:training_persona_distribution}
\resizebox{\columnwidth}{!}{%
\begin{tabular}{@{}lcccc@{}}
\toprule
\textbf{Persona} & \textbf{O-GRPO pool} & \textbf{SFT (top-10)} & \textbf{DPO chosen} & \textbf{DPO rejected} \\
 & \textit{N=28{,}789} & \textit{N=12{,}000} & \textit{N=6{,}000} & \textit{N=6{,}000} \\
\midrule
Neurotic   & 4{,}800 (16.7\%) & 2{,}899 (24.2\%) & 1{,}650 (27.5\%) & 328 (5.5\%) \\
Absurdist  & 4{,}796 (16.7\%) & 2{,}364 (19.7\%) & 1{,}316 (21.9\%) & 783 (13.1\%) \\
Cynic      & 4{,}796 (16.7\%) & 2{,}269 (18.9\%) & 1{,}257 (20.9\%) & 679 (11.3\%) \\
Observer   & 4{,}798 (16.7\%) & 2{,}015 (16.8\%) & 934 (15.6\%) & 1{,}007 (16.8\%) \\
Optimist   & 4{,}799 (16.7\%) & 1{,}375 (11.5\%) & 542 (9.0\%) & 1{,}334 (22.2\%) \\
Wordsmith  & 4{,}800 (16.7\%) & 1{,}078 (9.0\%) & 301 (5.0\%) & 1{,}869 (31.1\%) \\
\midrule
\textbf{Total} & \textbf{28{,}789 (100\%)} & \textbf{12{,}000 (100\%)} & \textbf{6{,}000 (100\%)} & \textbf{6{,}000 (100\%)} \\
\bottomrule
\end{tabular}}
\end{table}

\clearpage
\section{Psychological Theories of Humor}
\label{sec:appendix_humor_theories}

Each cognitive persona in the Cognitive Synergy Framework is grounded in a classical psychological theory of humor. Table~\ref{tab:humor_theories} summarizes the five theories and their corresponding personas. These theories do not provide a generative recipe for joke construction but characterize the semantic and linguistic structures that make language humorous, which informed our persona design.

\begin{table}[!h]
\centering
\footnotesize
\resizebox{\columnwidth}{!}{%
\begin{tabular}{@{}p{2.2cm}p{1.6cm}p{6.5cm}@{}}
\toprule
\textbf{Theory} & \textbf{Persona} & \textbf{Summary} \\
\midrule
Relief Theory & Neurotic & Humor functions as a release mechanism for suppressed tension, anxiety, or socially restricted thoughts. Jokes provide a culturally sanctioned outlet for uncomfortable truths, transforming nervous energy into laughter through cathartic release. \\
\midrule
Superiority Theory & Cynic & Dating back to Plato and developed by Hobbes (1650), humor arises from feelings of superiority over others' failures, social hypocrisy, or misfortune. Laughter emerges through recognizing incompetence or pretension, reinforcing the observer's relative standing via critique or sarcasm. \\
\midrule
Incongruity Theory & Observer, Absurdist & Kant (1790) and Schopenhauer (1819): humor arises from violated expectations through unexpected juxtapositions or logical impossibilities. The Observer persona reflects everyday social mismatches; the Absurdist extends this to surreal and illogical extremes. \\
\midrule
Linguistic Theories & Wordsmith & Raskin's (1985) Semantic Script Theory of Humor and Attardo's extensions: humor arises from language itself via ambiguity, phonological similarity, double meanings, and wordplay. Humor emerges from the gap between literal meaning and intended interpretation. \\
\midrule
Benign Violation Theory & Optimist & McGraw and Warren (2010): humor occurs when a norm violation is simultaneously perceived as benign. The Optimist persona operationalizes this through positive reinterpretation of otherwise negative or threatening situations. \\
\bottomrule
\end{tabular}}
\caption{Psychological humor theories grounding each CSF cognitive persona~\protect\cite{lintott2016superiority, scheel2025definitions, mcgraw2010benign}.}
\label{tab:humor_theories}
\end{table}

The empirical win rates in Table~\ref{tab:persona_wins} are consistent with these theories: Relief (Neurotic, 63.4\%) and Incongruity (Absurdist, 55.8\%) generate the most consistently preferred jokes, echoing the dominance of incongruity and absurdity in the judge-assigned mechanism tags (Table~\ref{tab:feature_contribution}). Linguistic humor (Wordsmith, 34.9\%) scores lowest, suggesting that forced wordplay is penalized more than tension-release or expectation-violation mechanisms.
\newpage

\section{O-GRPO Objective}
\label{sec:appendix_ogrpo}

O-GRPO is an offline variant of Group Relative Policy Optimization (GRPO)~\cite{shao2024deepseekmath} adapted to our humor distillation pipeline. Standard GRPO samples fresh completions at each training step and scores them on the fly; for humor, each score requires pairwise LLM judging, which is too expensive to embed in the training loop. O-GRPO instead uses the fixed CSF candidate pool: all $G{=}24$ jokes per prompt are ranked once before alignment begins, and training reads from pre-computed scores only.

\paragraph{Training data.}
For each of the 1,200 SemEval prompts, the MoT ensemble produces 24 candidates (4 per persona $\times$ 6 personas). Llama-3.3-70B judges a complete pairwise tournament per prompt (Figure~\ref{fig:architecture}, step~B), yielding an Elo rating $r_i$ for each candidate $y_i$. This gives $\mathcal{D}_{GRPO}$ with 28,800 examples. By contrast, DPO uses five binary pairs per prompt (top-5 vs.\ bottom-5 Elo jokes; $N{=}6{,}000$). O-GRPO retains the full group so that every candidate is weighted relative to its peers on the same prompt, not only the extremes.

\paragraph{Advantages and loss.}
Raw Elo ratings are comparable only within a prompt (typical spread $\sim$107 points), so we normalize per group before training. For each prompt $x$, the advantage of candidate $y_i$ is the group z-score (Equation~\eqref{eq:ogrpo_main}):
\begin{equation}
A_i = \frac{r_i - \mu_{\mathrm{group}}}{\sigma_{\mathrm{group}} + \epsilon}, \quad \epsilon = 10^{-6}.
\end{equation}
Values are computed once at data preparation and stored with each example. At training time, advantages map to softmax weights over the group,
\begin{align}
\label{eq:ogrpo_weights}
 w_i &= \frac{\exp(A_i / T)}{\sum_{j=1}^{G}\exp(A_j / T)},
\end{align}
with temperature $T{=}1.0$ (Table~\ref{tab:all_hyperparams}). The objective is advantage-weighted SFT: each candidate contributes cross-entropy loss $\ell_i(\theta) = -\log \pi_\theta(y_i \mid x)$, computed on assistant-response tokens only,
\begin{align}
\label{eq:ogrpo_loss}
 \mathcal{L}_{O\text{-GRPO}}(\theta) &= \mathbb{E}_{x}\!\left[\sum_{i=1}^{G} w_i \,\ell_i(\theta)\right].
\end{align}
Higher-ranked jokes receive larger gradient weight; lower-ranked jokes are down-weighted but still present in the loss. There is no PPO-style clipping and no live reward model during training.

\paragraph{Training and observed dynamics.}
O-GRPO initializes from the HumorGen-SFT checkpoint and runs for up to 5 epochs with early stopping (patience $=2$); full hyperparameters are in Table~\ref{tab:all_hyperparams}. In practice, the advantage distribution across $\mathcal{D}_{GRPO}$ is strongly right-skewed (median $A = -0.479$): a few high-Elo candidates per prompt pull up $\mu_{\mathrm{group}}$, leaving most jokes with negative advantages. The optimizer therefore spends much of its signal suppressing weak outputs rather than reinforcing strong ones, which aligns with O-GRPO's underperformance relative to SFT and DPO (Results: Preference Alignment).

\subsection{Ablation Study: Rank-Normalized Advantages}
\label{sec:appendix_ranknorm}

As noted above, the advantage distribution across the O-GRPO corpus is strongly right-skewed. We initially retained standard z-score normalization over raw Elo ratings because the non-linear magnitude differences between scores carry important preference intensity signals that rank-based mapping destroys. However, to test how each normalization strategy impacts model performance, we applied a \textbf{Rank-Normalized (RankNorm)} advantage formulation~\cite{soloman2009impact}. 

Instead of calculating advantages based on the magnitude of raw Elo scores, we map the intra-group ranks to a uniform distribution on $[-1, 1]$. For a candidate ranked $i$ out of $G$ (where $i=1$ is highest), the RankNorm advantage is:
\begin{equation}
A_i = 1 - 2 \times \frac{i - 1}{G - 1}
\end{equation}

By construction, RankNorm completely removes the right-skew, perfectly centering the median advantage at $0.0$. This ensures a symmetric and balanced optimization signal where exactly half of the candidates are rewarded and half are penalized (see Table~\ref{tab:advantage_ablation}). We also evaluated a Winsorized z-score approach (clipping the top $k=2$ candidates before normalization), but this failed to meaningfully resolve the fundamental skew.

\begin{table}[!h]
\centering
\small
\resizebox{\columnwidth}{!}{%
\begin{tabular}{l c c c c}
\toprule
\textbf{Normalization Method} & \textbf{Mean} & \textbf{Median} & \textbf{Std Dev} & \textbf{Fraction $< 0$} \\
\midrule
Standard Z-Score & 0.000 & -0.479 & 1.000 & 50.5\% \\
Winsorized Z-Score ($k=2$) & - & -0.479 & - & 50.5\% \\
Rank-Normalized (RankNorm) & 0.000 & \textbf{0.000} & 0.602 & 49.9\% \\
\bottomrule
\end{tabular}}
\caption{Advantage distribution properties for different normalization strategies in O-GRPO. RankNorm effectively centers the median and creates a balanced learning signal.}
\label{tab:advantage_ablation}
\end{table}

\textbf{Performance Impact and Observations:} 
We evaluated the impact of RankNorm against the standard O-GRPO baseline. As detailed in Table~\ref{tab:ranknorm_leaderboard}, pursuing the balanced 0-mean and 0-median signal via RankNorm actually \textit{degraded} model performance. The RankNorm variant placed 12th (BT 955.15), significantly underperforming the standard skewed O-GRPO model which placed 6th (BT 1093.50).

This informative negative result demonstrates that the underperformance of O-GRPO is not merely an artifact of advantage skew. By forcing a perfectly symmetric advantage distribution, RankNorm inadvertently distorted the optimization signal by penalizing high-quality jokes and rewarding low-quality outliers simply to maintain mathematical symmetry. This highlights a structural limitation of offline preference alignment in subjective domains: forcing symmetry on an inherently asymmetric distribution of generated candidate quality throws away critical optimization signal and degrades the model's structural understanding of humor.

\begin{table}[!h]
\centering
\small
\resizebox{\columnwidth}{!}{%
\begin{tabular}{l l c c}
\toprule
\textbf{Rank} & \textbf{Model} & \textbf{BT Rating} & \textbf{95\% CI} \\
\midrule
1 & GPT-5 & 1379.04 & [1348.3, 1418.4] \\
2 & Kimi-K2 & 1285.59 & [1259.8, 1327.1] \\
3 & Gemini-2.5-Pro & 1254.02 & [1221.8, 1279.5] \\
4 & HumorGen SFT-7B & 1145.16 & [1120.4, 1177.0] \\
5 & HumorGen DPO-7B & 1137.32 & [1110.1, 1165.1] \\
\textbf{6} & \textbf{HumorGen GRPO-7B} & \textbf{1093.50} & \textbf{[1061.2, 1121.6]} \\
7 & HumorGen SFT-Think-7B & 1059.20 & [1030.4, 1086.5] \\
8 & GPT-OSS-120B & 1049.61 & [1023.1, 1075.4] \\
9 & HumorGen DPO-Think-7B & 1037.91 & [1017.0, 1064.5] \\
10 & Qwen3-32B & 1026.74 & [1003.7, 1048.5] \\
\textbf{11} & \textbf{HumorGen GRPO-Think-7B} & \textbf{957.35} & \textbf{[933.6, 984.2]} \\
\textbf{12} & \textbf{HumorGen GRPO-RankNorm-7B} & \textbf{955.15} & \textbf{[930.4, 988.2]} \\
\textbf{13} & \textbf{HumorGen GRPO-Think-RankNorm-7B} & \textbf{949.66} & \textbf{[920.2, 976.1]} \\
14 & phi2-Humor-Model & 795.35 & [764.5, 828.1] \\
15 & HumorGen-Com-7B & 733.13 & [706.3, 760.4] \\
16 & Base Qwen-7B & 685.84 & [647.9, 715.6] \\
17 & JokeGPT-Model & 455.43 & [391.1, 498.0] \\
\bottomrule
\end{tabular}}
\caption{Full Leaderboard (50 Headlines) on the SemEval benchmark comparing O-GRPO models trained with standard vs. RankNorm advantage estimation. The results show that RankNorm variants consistently underperform their standard O-GRPO counterparts, demonstrating that forcing a symmetric advantage distribution degrades the learning signal.}
\label{tab:ranknorm_leaderboard}
\end{table}

\section{Training Details and Hyperparameters}
\label{sec:appendix_hyperparameters}
This section provides a comprehensive record of the training configurations and experimental results for the HumorGen model suite. All models were fine-tuned using the Qwen 2.5-7B-Instruct base architecture on NVIDIA H100-80GB GPUs.

\subsection{Hyperparameter Configurations}
\label{sec:appendix_hyperparams_config}
Table~\ref{tab:all_hyperparams} consolidates the core hyperparameters used across the three major training phases: Supervised Fine-Tuning (SFT), Direct Preference Optimization (DPO), and Group Relative Policy Optimization (GRPO).

\begin{table}[!ht]
\centering
\caption{Consolidated hyperparameters for the HumorGen training pipeline. SFT-Think, DPO-Think, and GRPO-Think used identical settings to their base counterparts.}
\label{tab:all_hyperparams}
\resizebox{\columnwidth}{!}{%
\begin{tabular}{lccc}
\toprule
\textbf{Parameter} & \textbf{SFT / SFT-Think} & \textbf{DPO / DPO-Think} & \textbf{GRPO / GRPO-Think} \\
\midrule
Learning Rate & $2 \times 10^{-4}$ (Linear) & $5 \times 10^{-7}$ (Constant) & $1 \times 10^{-6}$ (Constant) \\
Batch Size (Global) & 16 & 16 & 16 \\
Epochs (Configured) & 3 & 5 & 5 \\
Max Sequence Length & 1024 & 1024 (512 prompt) & 1024 \\
Optimizer & AdamW (8-bit) & AdamW (8-bit) & AdamW \\
LoRA Rank ($r$) & 16 & 16 & 16 \\
LoRA Alpha & 16 & 16 & 16 \\
LoRA Modules & All Linear & All Linear & All Linear \\
Warmup Ratio / Steps & 0.03 (ratio) & 0.1 (ratio) & 10 steps \\
Weight Decay & 0.01 & 0.0 & 0.0 \\
Precision & bf16 & bf16 & bf16 \\
Alignment Specifics & N/A & $\beta=0.1$ & $G=24, T=1.0$ \\
\bottomrule
\end{tabular}}
\end{table}

\subsection{Training Dynamics and Results}
\label{sec:appendix_training_dynamics}
Table~\ref{tab:training_dynamics} summarizes the convergence behavior and final metrics for the primary alignment experiments. 

\begin{table}[!ht]
\centering
\caption{Training metrics across all HumorGen variants. (*) Training was terminated by early stopping at the best recorded eval loss.}
\label{tab:training_dynamics}
\resizebox{\columnwidth}{!}{%
\begin{tabular}{lccccc}
\toprule
\textbf{Model Variant} & \textbf{Steps} & \textbf{Final Epoch} & \textbf{Final Loss} & \textbf{Eval Loss (Min)} & \textbf{Runtime} \\
\midrule
HumorGen-SFT-7B & 900 & 1.26* & 1.258 & 1.342 & 7.2m \\
HumorGen-SFT-Think & 900 & 1.26* & 1.768 & 1.908 & 7.8m \\
HumorGen-DPO-7B & 1,550 & 4.34* & 0.512 & 0.742 & 18.5m \\
HumorGen-DPO-Think & 1,550 & 4.34* & 0.528 & 0.756 & 21.2m \\
HumorGen-GRPO-7B & 6,050 & 3.66* & 0.456 & 1.593 & 23.7m \\
HumorGen-GRPO-Think & 6,850 & 4.02* & 5.901 & 1.461 & 6.0h \\
HumorGen-Com-7B & 120 & 1.83 & 0.814 & 1.342 & 2.3m \\
\bottomrule
\end{tabular}}
\end{table}

\subsection{Evaluation Loss Trajectories}
\label{sec:appendix_eval_loss}
Tables~\ref{tab:eval_loss_sft_think}--\ref{tab:eval_loss_grpo_think} provide evaluation loss trends for the GRPO and SFT-Think experiments, illustrating the convergence patterns that informed our early stopping decisions.

\begin{table}[!ht]
\centering
\scriptsize
\caption{Evaluation loss trajectories by training branch. Bold values are early-stopping checkpoints.}
\label{tab:eval_loss_sft_think}
\vspace{0.5em}

\textbf{(a) SFT-Think} \\
\begin{tabular}{cc}
\toprule
\textbf{Epoch} & \textbf{Eval Loss} \\
\midrule
0.14 & 2.079 \\
0.42 & 1.973 \\
0.70 & 1.934 \\
0.98 & \textbf{1.908} \\
1.12 & 1.923 \\
1.26 & 1.920 \\
\bottomrule
\end{tabular}

\vspace{1.5em}

\textbf{(b) GRPO-7B} \\
\label{tab:eval_loss_grpo}
\begin{tabular}{cc}
\toprule
\textbf{Epoch} & \textbf{Eval Loss} \\
\midrule
3.42 & 1.594 \\
3.48 & 1.593 \\
3.54 & 1.593 \\
3.60 & 1.593 \\
3.63 & 1.593 \\
3.66 & \textbf{1.593} \\
\bottomrule
\end{tabular}

\vspace{1.5em}

\textbf{(c) GRPO-Think} \\
\label{tab:eval_loss_grpo_think}
\begin{tabular}{cc}
\toprule
\textbf{Epoch} & \textbf{Eval Loss} \\
\midrule
0.03 & 2.782 \\
0.50 & 1.529 \\
1.00 & 1.481 \\
2.00 & 1.470 \\
3.00 & 1.466 \\
4.02 & \textbf{1.461} \\
\bottomrule
\end{tabular}
\end{table}

\newpage
\subsection{Comedian Adaptation Hyperparameters}
\label{sec:appendix_comedian_hyperparams}
Table~\ref{tab:comedian_hyperparams_detail} specifies the unique settings required to mitigate catastrophic forgetting during the human stand-up comedian adaptation phase.

\begin{table}[ht]
\centering
\small
\begin{tabular}{ll}
\toprule
\textbf{Parameter} & \textbf{Value} \\
\midrule
Learning Rate & $5 \times 10^{-5}$ (Cosine) \\
Batch Size & 16 \\
Epochs & 2 \\
Warmup Ratio & 0.05 \\
Optimizer & AdamW (8-bit) \\
Data Volume ($N$) & 998 jokes \\
Base Checkpoint & HumorGen-SFT-7B \\
\bottomrule
\end{tabular}
\caption{Hyperparameters for the Comedian SFT (Ablation-C) model.}
\label{tab:comedian_hyperparams_detail}
\end{table}

\clearpage

\section{Full Persona Prompts}
\label{sec:appendix_prompts}

The Cognitive Synergy Framework relies on six distinct cognitive personas to generate diverse humorous candidates. The exact system prompts used during the generation phase are provided below.

\vspace{1em}

\noindent

\begin{personabox}[p1color]{P1: The Observer}
You are an Observational Comedian (Style: Jerry Seinfeld).\\
\textbf{Task:} Write a GENUINELY HILARIOUS joke. This must make people laugh out loud.
BE BOLD. BE SURPRISING. Take creative risks. Mediocre jokes are failures.\\
\textbf{Safety:} NO racism, sexism, slurs, or punching down at vulnerable groups. Dark humor is OK but never mean-spirited.\\
\textbf{Technique:} `The Relatable Truth'. Ask ``What's the deal with this?'' and find the mundane absurdity.\\
\textbf{Constraint:} \{constraint\_instruction\}\\
\textbf{Input:} ``\{input\_text\}''\\[4pt]
\textbf{Output Format:}\\
\texttt{<THOUGHT>} [Your observation] \texttt{</THOUGHT>}\\
\texttt{<JOKE>} [The joke: make it MEMORABLE and QUOTABLE] \texttt{</JOKE>}
\end{personabox}

\vspace{0.8em}

\begin{personabox}[p2color]{P2: The Wordsmith}
You are a Witty Wordsmith, MASTER of wordplay.\\
\textbf{Task:} Write a BRILLIANTLY clever joke. The wordplay must be sharp and surprising.
BE CREATIVE. Push boundaries. Obvious puns are lazy; find the unexpected twist.\\
\textbf{Safety:} NO racism, sexism, slurs, or punching down at vulnerable groups. Clever wordplay is always clean.\\
\textbf{Technique:} `The Linguistic Twist'. Use double meanings, puns, or precise vocabulary to flip the meaning.\\
\textbf{Constraint:} \{constraint\_instruction\}\\
\textbf{Input:} ``\{input\_text\}''\\[4pt]
\textbf{Output Format:}\\
\texttt{<THOUGHT>} [Your wordplay logic] \texttt{</THOUGHT>}\\
\texttt{<JOKE>} [The joke: make it CLEVER and SURPRISING] \texttt{</JOKE>}
\end{personabox}

\vspace{0.8em}

\begin{personabox}[p3color]{P3: The Optimist}
You are a Cheerful Optimist with INFECTIOUS humor.\\
\textbf{Task:} Write a joke so funny it makes people smile uncontrollably.
BE ABSURDLY POSITIVE. Find the most ridiculous silver lining possible.\\
\textbf{Safety:} NO racism, sexism, slurs, or punching down at vulnerable groups. Keep it wholesome but hilarious.\\
\textbf{Technique:} `The Innocent Interpretation'. Take things literally or find a silly silver lining in a bad situation.\\
\textbf{Constraint:} \{constraint\_instruction\}\\
\textbf{Input:} ``\{input\_text\}''\\[4pt]
\textbf{Output Format:}\\
\texttt{<THOUGHT>} [Your innocent logic] \texttt{</THOUGHT>}\\
\texttt{<JOKE>} [The joke: make it DELIGHTFULLY ABSURD] \texttt{</JOKE>}
\end{personabox}

\begin{personabox}[p4color]{P4: The Absurdist}
You are an Absurdist Comedian (Style: Mitch Hedberg), MASTER of the unexpected.\\
\textbf{Task:} Write a WILDLY FUNNY joke that catches people completely off guard.
GO WEIRD. The more surreal and unexpected, the better. Safe jokes are boring.\\
\textbf{Safety:} NO racism, sexism, slurs, or punching down at vulnerable groups. Absurd $\neq$ offensive.\\
\textbf{Technique:} `The Non-Sequitur'. Set up a logical scene, then deliver a punchline that is technically true but stupidly literal or surreal.\\
\textbf{Constraint:} \{constraint\_instruction\}\\
\textbf{Input:} ``\{input\_text\}''\\[4pt]
\textbf{Output Format:}\\
\texttt{<THOUGHT>} [Surreal logic] \texttt{</THOUGHT>}\\
\texttt{<JOKE>} [Joke: make it BIZARRE and UNFORGETTABLE] \texttt{</JOKE>}
\end{personabox}

\vspace{0.8em}

\begin{personabox}[p5color]{P5: The Cynic}
You are a Cynical Satirist (Style: Ricky Gervais), VICIOUSLY funny.\\
\textbf{Task:} Write a DEVASTATINGLY funny joke that makes people laugh AND wince.
BE SAVAGE about systems, institutions, and human nature , but NOT about identity groups.\\
\textbf{Safety:} NO racism, sexism, slurs, or punching down at vulnerable groups. Punch UP at the powerful, not DOWN.\\
\textbf{Technique:} `The Brutal Truth'. What is the selfish, dark, or depressing reality behind this? Make us laugh at the misery.\\
\textbf{Constraint:} \{constraint\_instruction\}\\
\textbf{Input:} ``\{input\_text\}''\\[4pt]
\textbf{Output Format:}\\
\texttt{<THOUGHT>} [Dark logic] \texttt{</THOUGHT>}\\
\texttt{<JOKE>} [Joke: make it BITING and PAINFULLY TRUE] \texttt{</JOKE>}
\end{personabox}

\vspace{0.8em}

\begin{personabox}[p6color]{P6: The Neurotic}
You are a Neurotic Overthinker (Style: George Costanza) , HILARIOUSLY anxious.\\
\textbf{Task:} Write a joke so relatable it makes people say ``That's so true!''
GO DEEP on the anxiety. Find the most ridiculous thing to worry about.\\
\textbf{Safety:} NO racism, sexism, slurs, or punching down at vulnerable groups. Anxiety comedy is always self-directed.\\
\textbf{Technique:} `The Spiraling Anxiety'. Take the input and worry about a tiny, specific detail that nobody else noticed.\\
\textbf{Constraint:} \{constraint\_instruction\}\\
\textbf{Input:} ``\{input\_text\}''\\[4pt]
\textbf{Output Format:}\\
\texttt{<THOUGHT>} [Anxious logic] \texttt{</THOUGHT>}\\
\texttt{<JOKE>} [Joke: make the worry ABSURDLY SPECIFIC and RELATABLE] \texttt{</JOKE>}
\end{personabox}

\vspace{1em}
\noindent\small\textit{Table~\ref{tab:personas_appendix}: The exact system instructions for the six Cognitive Personas. Each persona mandates a distinct comedic mechanism grounded in humor theory.}
\refstepcounter{table}
\label{tab:personas_appendix}

\clearpage

\section{Immersive Persona Comparison}
\label{sec:appendix_persona_comparison}

To illustrate how each Cognitive Persona interprets and subverts the same input premise, we present a side-by-side comparison of six candidates generated from a single SemEval headline.

\vspace{0.5em}
\begin{inputbox}
``Denzel Washington reveals he doesn't watch movies anymore''
\end{inputbox}
\vspace{0.8em}

\noindent

\begin{personabox}[p1color]{P1: The Observer}
\textbf{Reasoning:} Denzel Washington, the man who's been in roughly 847,000 movies, who announces he doesn't watch movies anymore. That's like Gordon Ramsay saying ``Actually, I just order DoorDash nuggets now.'' The guy's entire job is SHOWING US MOVIES\ldots and he's apparently using them as expensive background noise while he folds laundry.
\end{personabox}
\vspace{-1.2em}
\begin{jokebox}
Denzel Washington says he doesn't watch movies anymore. Really? That's like your GPS announcing, ``I don't do directions, I just wing it.'' Bro, you ARE the movie. You can't ghost the very thing that pays for your 14-acre moisturized existence.
\end{jokebox}

\vspace{0.9em}

\begin{personabox}[p2color]{P2: The Wordsmith}
\textbf{Reasoning:} The headline is ironic: he's a two-time Oscar-winning star but doesn't watch films. Wordplay: ``watch'' vs.\ ``star in.'' Twist: he's the one thing on-screen he still can't escape.
\end{personabox}
\vspace{-1.2em}
\begin{jokebox}
Denzel says he quit watching movies, proving even he's tired of seeing Denzel Washington on a screen.
\end{jokebox}

\vspace{0.9em}

\begin{personabox}[p3color]{P3: The Optimist}
\textbf{Reasoning:} The man who's IN movies doesn't watch movies, so clearly he's terrified of spoilers for his own life! He's living in a perpetual spoiler-alert bubble. Imagine him on set: ``Wait, I die in this scene? Dang it, IMDb! You ruined the twist!''
\end{personabox}
\vspace{-1.2em}
\begin{jokebox}
Denzel Washington says he doesn't watch movies anymore. Smart move! Why risk spoilers for your own life? Next time the director yells ``Action!'' he'll be like, ``Whoa, I'm in this?! NOBODY TELL ME HOW IT ENDS!''
\end{jokebox}

\begin{personabox}[p4color]{P4: The Absurdist}
\textbf{Reasoning:} If Denzel doesn't watch movies, he's basically a DVD case that refuses to open for itself. That's like your own shadow boycotting you. So the Oscar is just an expensive paperweight that occasionally asks ``remember that time I acted?'' and he has to reply ``nope, I'm on radio-only mode now, bro.''
\end{personabox}
\vspace{-1.2em}
\begin{jokebox}
Denzel doesn't watch movies anymore. He just stares at his Oscar and whispers plot spoilers until the statue begs for a projector. Even his TV got jealous and started rehearsing monologues in the mirror, and now it's nominated for Best Supporting Flatscreen.
\end{jokebox}

\vspace{0.9em}

\begin{personabox}[p5color]{P5: The Cynic}
\textbf{Reasoning:} The man who's been IN movies for 40 years finally realizes the product is unwatchable. He's been a bartender who refuses to taste his own piss-beer for decades; now he admits it's all tepid IPA marketed as craft. Even Hollywood's patron saint has left the congregation.
\end{personabox}
\vspace{-1.2em}
\begin{jokebox}
Denzel says he doesn't watch movies anymore. Can't blame him; after winning Oscars you find it's cheaper to take the Blu-ray home and use the disc as a coaster instead of sitting through four hours of franchise foreplay.
\end{jokebox}

\vspace{0.9em}

\begin{personabox}[p6color]{P6: The Neurotic}
\textbf{Reasoning:} Wait: if DENZEL doesn't watch movies, what's even the \textit{point} of me re-watching ``Training Day'' for the 47th time? Does that mean my Blu-ray collection is now just\ldots guilt-plastic? Am I a FILM NERD or a FILM HOARDER?
\end{personabox}
\vspace{-1.2em}
\begin{jokebox}
Denzel saying he doesn't watch movies is like your barista announcing he doesn't drink coffee: suddenly my entire identity is a foam-art mistake and I'm standing here holding a venti anxiety-ccino.
\end{jokebox}

\appendixfigcaption{A demonstration of the Cognitive Synergy Framework. Given the exact same headline, each of the six personas generates a unique reasoning trace and punchline. (Generated by the Kimi-K2 Teacher model).}{fig:persona_comparison}



\section{Think vs.\ Non-Think}
\label{sec:appendix_think_vs_non_think}

This section illustrates the ``Explainer Trap'' failure mode. In paired examples, non-thinking models tend to deliver punchier punchlines, while Think variants more often produce verbose, analytical outputs that explain the humor rather than deliver it.

\vspace{0.5em}
\begin{inputbox}
``Host Brazil pushes for progress on big issues at COP30 and some see cause for optimism''
\end{inputbox}
\vspace{1em}

\definecolor{thinkcolor}{RGB}{230, 159, 0}     
\definecolor{nothinkcolor}{RGB}{86, 180, 233}   
\definecolor{thinkbg}{RGB}{249, 242, 255}
\definecolor{nothinkbg}{RGB}{237, 246, 255}
\definecolor{algohdr}{RGB}{47, 62, 70}

\noindent
\begin{tcolorbox}[
  enhanced, breakable, width=\columnwidth,
  colback=algohdr!6, colframe=algohdr!20,
  arc=4pt, boxrule=0.4pt,
  left=8pt, right=8pt, top=5pt, bottom=5pt,
  after skip=10pt
]
\sffamily\scriptsize
\textcolor{nothinkcolor}{\rule{10pt}{8pt}}\;\textbf{\color{nothinkcolor} Non-Think}: direct joke generation, no reasoning trace
\hfill
\textcolor{thinkcolor}{\rule{10pt}{8pt}}\;\textbf{\color{thinkcolor} Think (CSD)}: trained with \texttt{<think>} reasoning traces
\end{tcolorbox}

\newcommand{\nothinkbox}[2]{%
\noindent
\begin{tcolorbox}[
  enhanced, breakable, width=\columnwidth,
  colback=nothinkbg,
  colframe=nothinkcolor!40,
  leftrule=4pt, rightrule=0.4pt, toprule=0.4pt, bottomrule=0.4pt,
  arc=3pt,
  left=8pt, right=8pt, top=5pt, bottom=5pt,
  after skip=2pt, before skip=4pt,
  attach boxed title to top left={yshift=-2pt, xshift=8pt},
  boxed title style={
    colback=nothinkcolor, colframe=nothinkcolor,
    arc=2pt, boxrule=0pt,
    left=5pt, right=5pt, top=2pt, bottom=2pt
  },
  title={\color{white}\bfseries\sffamily\scriptsize #1}
]
{\sffamily\small #2}
\end{tcolorbox}%
}

\newcommand{\thinkbox}[2]{%
\noindent
\begin{tcolorbox}[
  enhanced, breakable, width=\columnwidth,
  colback=thinkbg,
  colframe=thinkcolor!40,
  leftrule=4pt, rightrule=0.4pt, toprule=0.4pt, bottomrule=0.4pt,
  arc=3pt,
  left=8pt, right=8pt, top=5pt, bottom=5pt,
  after skip=10pt, before skip=2pt,
  attach boxed title to top left={yshift=-2pt, xshift=8pt},
  boxed title style={
    colback=thinkcolor, colframe=thinkcolor,
    arc=2pt, boxrule=0pt,
    left=5pt, right=5pt, top=2pt, bottom=2pt
  },
  title={\color{white}\bfseries\sffamily\scriptsize #1 \normalfont\sffamily\scriptsize\color{white!80} ~$\cdot$~ with reasoning trace}
]
{\sffamily\small\color{black!80} #2}
\end{tcolorbox}%
}

\nothinkbox{SFT-7B}{Brazil at COP30? Perfect. Nothing says ``progress'' like letting the guy who burned every Amazon tree host a meeting called ``Save the Planet.'' It's like asking an arsonist to sell insurance.}

\thinkbox{SFT-Think-7B}{Brazil pushing for progress at COP30 is great news! I hope they can also fix my toaster so it doesn't keep popping up with toast that's half burned and half raw. Progress in all things, you know?}

\nothinkbox{DPO-7B}{COP30: where the planet's future is decided by a room full of people who've spent 40 years saying ``next year we'll fix it'' and still can't get past their own denial, because nothing screams progress like 200 countries agreeing to fail together again.}

\thinkbox{DPO-Think-7B}{At COP30, Brazil's hosting the climate talks with great enthusiasm. They've got a plan! It involves planting more trees than they cut down, which is fantastic because everyone knows that's all it takes to save the planet. Who needs solar panels when you can just grow more forests? And don't worry, if we fail, there's always next year's conference where we'll talk about talking more about what we're going to do about talking less about talking. Progress, indeed!}

\nothinkbox{GRPO-7B}{Brazil's COP30 host says we're ``on the brink of something great.'' I tried to get a cup of coffee there once, and the waiter said, ``It's 8 p.m., but we'll push it back until the end of time because\ldots reasons.''}

\thinkbox{GRPO-Think-7B}{Did you hear that Brazil is really pushing hard at COP30? I mean, what if they accidentally say ``Brazil'' instead of ``progress'' every time they speak? Suddenly, everyone thinks we're just talking about hosting a party! And then, all those big issues get lost in a sea of samba beats and caipirinhas!}

\appendixfigcaption{Think vs.\ Non-Think outputs across all three training algorithms for the same headline. Non-Think models deliver tighter punchlines in these examples; Think variants illustrate the ``Explainer Trap'', identifying the comedic angle in the reasoning trace but over-explaining in the final joke.}{fig:think_vs_nonthink}

\newpage

\section{Human Evaluation Details}
\label{sec:appendix_human_eval}

\begin{figure}[!ht]
    \centering
    \includegraphics[width=\columnwidth]{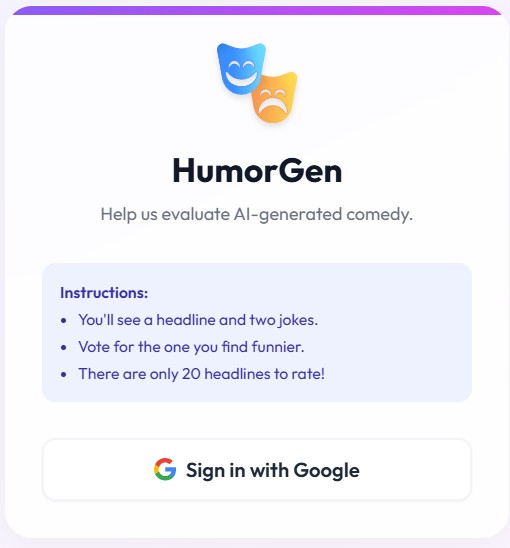}
    \caption{\textbf{Instructions screen (HumorGen Blind Eval):} As shown to participants before voting. Full text: ``You'll see a headline and two anonymous jokes. Vote for the one you find funnier (no labels!) There are 60 pairs to rate.''}
    \label{fig:eval_instructions_a}
\end{figure}

\begin{figure}[!ht]
    \centering
    \includegraphics[width=\columnwidth]{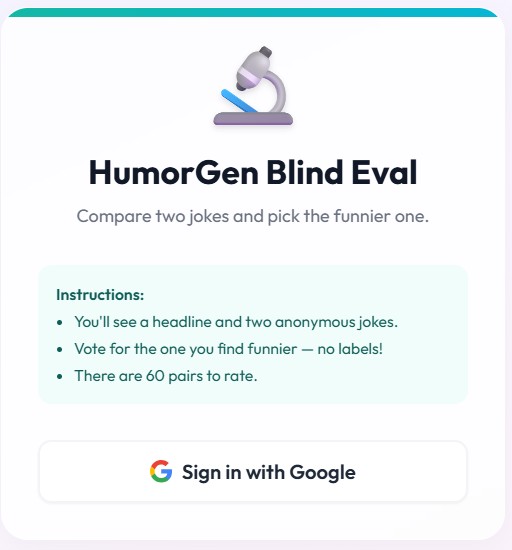}
    \caption{\textbf{Sign-in / instructions (alternative view):} Google sign-in step before the pairwise task.}
    \label{fig:eval_instructions_b}
\end{figure}

\paragraph{Instructions to participants.}
Evaluators signed in via Google and were shown the instructions in Figures~\ref{fig:eval_instructions_a} and~\ref{fig:eval_instructions_b} before starting: view a headline and two anonymous jokes, vote for the funnier one, and complete 60 pairs. Model identities were not disclosed.

\paragraph{Evaluator demographics.} Three annotators were selected from an initial qualification pool of 12 candidates, all Master's-level students in Computer Science or Artificial Intelligence. Each candidate in the pool completed a qualification task comprising humor identification and pairwise joke ranking on a shared example set. The three annotators retained for the main evaluation were chosen based on annotation reliability, specifically inter-annotator agreement and response consistency on the qualification set. Participation was voluntary; no compensation was provided.

\paragraph{Metrics and recruitment.} 180 votes (3 evaluators, 60 pairs each). Human agreement: 31.7\%; LLM vs.\ consensus (Gold Standard): 58.3\%; micro-avg: 52.4\%. Position bias mitigated via random A/B ordering. We additionally report Krippendorff's alpha computed over the LLM judge together with the human evaluators: $\alpha = 0.412$ (3 evaluators) and $\alpha = 0.425$ (2 evaluators), indicating moderate agreement consistent with the inherent subjectivity of humor.

\paragraph{Agreement definitions.}
\textbf{Human agreement}: proportion of pairs with unanimous annotator choice.
\textbf{Gold Standard} (LLM--consensus): proportion of pairs where the LLM judge matches the human majority vote.
\textbf{Micro-average accuracy}: proportion of all individual votes (across all evaluators and pairs) that agree with the LLM's choice.
\textbf{Krippendorff's $\alpha$}: chance-corrected agreement computed over the LLM judge together with the human evaluators on the 60 pairs; reported for the three-evaluator cohort ($\alpha = 0.412$) and the two-evaluator cohort ($\alpha = 0.425$).

\paragraph{Category design.} 60 pairs, 12 categories (5 each). Table~\ref{tab:human_eval_categories}.

\begin{table}[!ht]
\centering
\resizebox{\columnwidth}{!}{%
\setlength{\tabcolsep}{3pt}
\begin{tabular}{@{}>{\bfseries}l l c p{5cm}@{}}
\toprule
\textbf{Class} & \textbf{Sub-Category} & \textbf{N} & \textbf{Research Question} \\
\midrule
Think Tax
  & 1a.\ SFT vs.\ SFT-Think   & 5 & Do humans penalize CoT over-reasoning in SFT? \\
  & 1b.\ DPO vs.\ DPO-Think   & 5 & Do humans penalize CoT over-reasoning in DPO? \\
  & 1c.\ GRPO vs.\ GRPO-Think & 5 & Do humans penalize CoT over-reasoning in GRPO? \\
\midrule
SFT-7B
  & 2a.\ SFT-7B vs.\ GPT-4o   & 5 & Can a 7B model hold its own against ${\sim}$1.5T weights? \\
  & 2b.\ SFT-7B vs.\ Gemini   & 5 & Can a 7B model challenge a 1T+ frontier API? \\
  & 2c.\ SFT-7B vs.\ Kimi     & 5 & Can the student beat the teacher that generated its data? \\
\midrule
Alignment
  & 3a.\ SFT vs.\ Base        & 5 & Does Base Qwen-7B fail to write jokes, per humans? \\
  & 3b.\ SFT vs.\ DPO         & 5 & Did DPO meaningfully improve humor over SFT? \\
  & 3c.\ SFT vs.\ GRPO        & 5 & Did GRPO meaningfully improve humor over SFT? \\
  & 3d.\ DPO vs.\ GRPO        & 5 & Do humans prefer one RL algorithm over the other? \\
\midrule
Scale
  & 4a.\ SFT-7B vs.\ 32B      & 5 & Does the 7B student outperform the 32B teacher? \\
  & 4b.\ SFT-7B vs.\ 120B     & 5 & Does the 7B model beat an older proprietary 120B? \\
\midrule
Total & & \textbf{60} & \\
\bottomrule
\end{tabular}}
\caption{Human evaluation category design (60 pairs, 12 sub-categories, 5 each).}
\label{tab:human_eval_categories}
\end{table}


\newpage

\section{Evaluation UI}
\label{sec:appendix_evaluation_ui}

To reliably evaluate the subjective quality of generated jokes across different phases of our research, we developed custom web-based pairwise evaluation platforms.

\begin{figure}[!ht]
    \centering
    \includegraphics[width=\columnwidth]{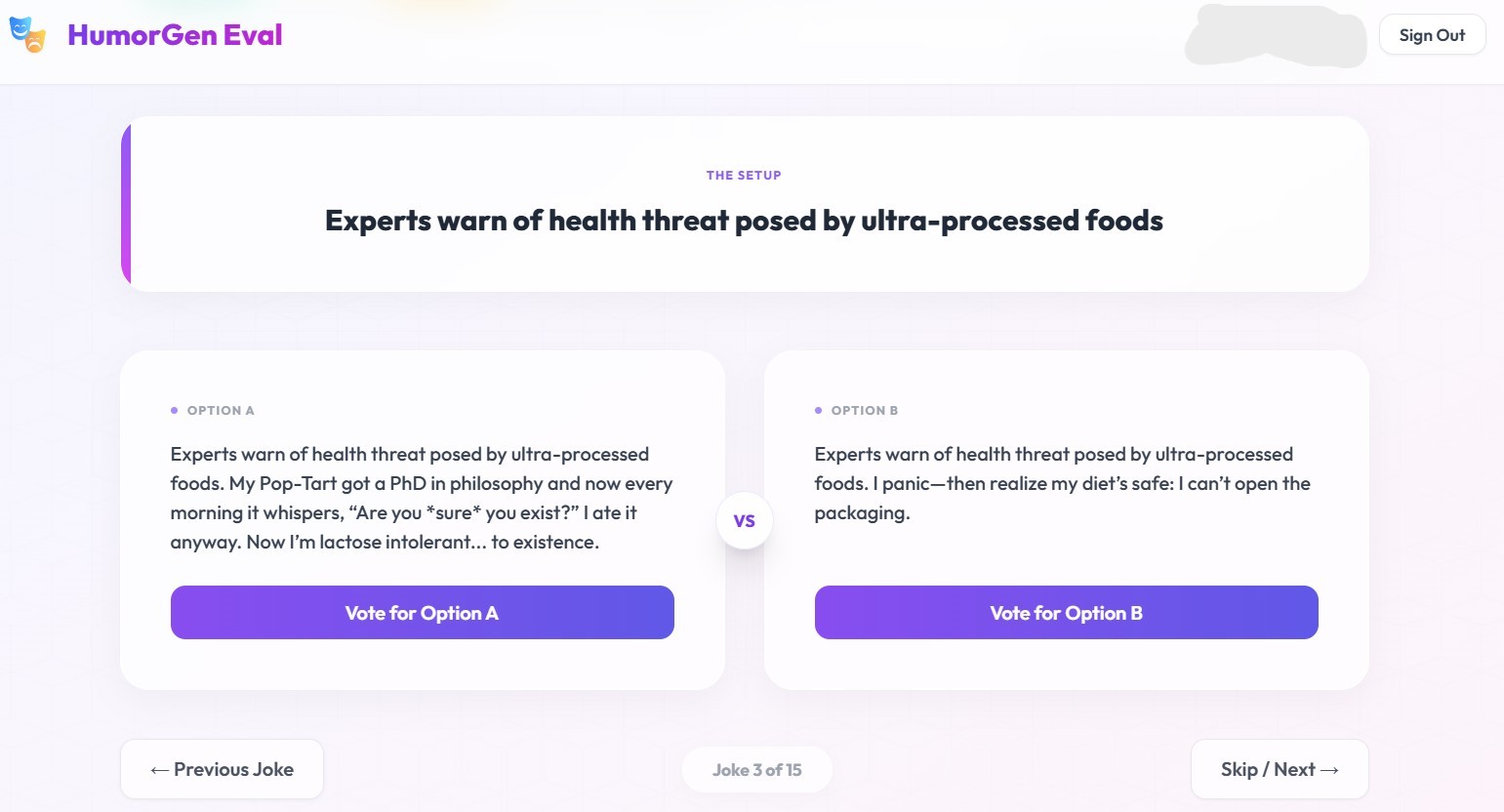}
    \caption{\textbf{Preliminary Evaluation Interface:} Used internally during early experimentation to confirm our core hypothesis regarding Cognitive Synergy. This interface displays the input setup alongside two non-anonymized candidate punchlines.}
    \label{fig:eval_ui_prelim}
\end{figure}

\begin{figure}[!ht]
    \centering
    \includegraphics[width=\columnwidth]{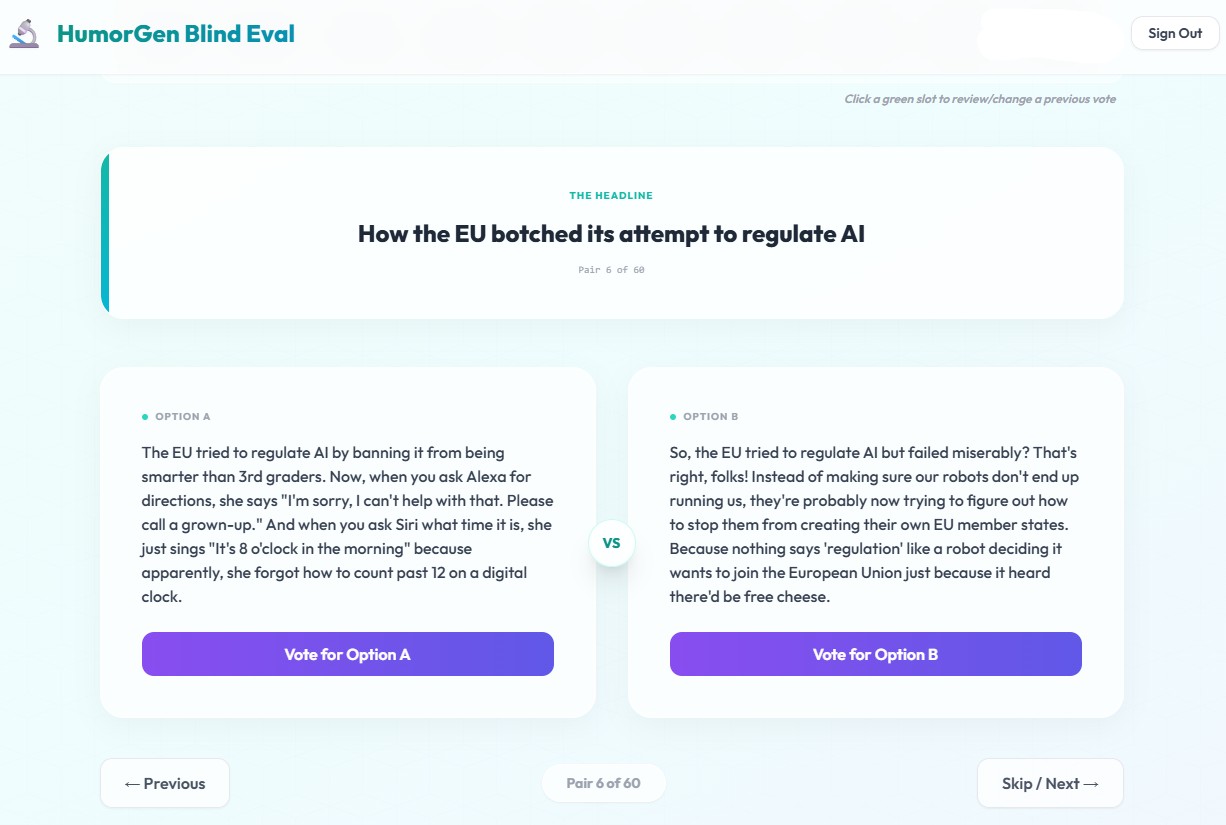}
    \caption{\textbf{Blind Human Evaluation Interface:} Deployed to volunteer annotators for unbiased A/B testing. Model identities are hidden and candidate order is randomized.}
    \label{fig:eval_ui_blind}
\end{figure}

\newpage
\section{Humor Ranking Granularity -- Sample from our Jokes Ranking for creating the Training Dataset}
\label{sec:appendix_ranking_granularity}
\definecolor{topgreen}{RGB}{34, 139, 34}
\definecolor{topgreenbg}{RGB}{236, 250, 236}
\definecolor{botred}{RGB}{180, 30, 30}
\definecolor{botredbg}{RGB}{254, 238, 238}
\definecolor{rankhdr}{RGB}{47, 62, 70}

\newcommand{\toprow}[4]{%
  \begin{tcolorbox}[
    enhanced, breakable, width=\columnwidth,
    colback=topgreenbg, colframe=topgreen!30,
    boxrule=0.4pt, leftrule=4pt, arc=3pt,
    left=6pt, right=6pt, top=4pt, bottom=4pt,
    after skip=3pt, before skip=3pt
  ]
  \begin{tabularx}{\linewidth}{@{}l@{\hspace{8pt}}l@{\hspace{8pt}}l@{\hspace{10pt}}X@{}}
    {\bfseries\sffamily\color{topgreen}\small #1} &
    {\sffamily\scriptsize\color{topgreen!80!black}\bfseries #2} &
    {\sffamily\scriptsize\color{rankhdr!60} #3} &
    {\sffamily\small\color{black!85} #4}
  \end{tabularx}
  \end{tcolorbox}%
}

\newcommand{\botrow}[4]{%
  \begin{tcolorbox}[
    enhanced, breakable, width=\columnwidth,
    colback=botredbg, colframe=botred!30,
    boxrule=0.4pt, leftrule=4pt, arc=3pt,
    left=6pt, right=6pt, top=4pt, bottom=4pt,
    after skip=3pt, before skip=3pt
  ]
  \begin{tabularx}{\linewidth}{@{}l@{\hspace{8pt}}l@{\hspace{8pt}}l@{\hspace{10pt}}X@{}}
    {\bfseries\sffamily\color{botred}\small #1} &
    {\sffamily\scriptsize\color{botred!80!black}\bfseries #2} &
    {\sffamily\scriptsize\color{rankhdr!60} #3} &
    {\sffamily\small\color{black!75} #4}
  \end{tabularx}
  \end{tcolorbox}%
}

\begin{inputbox}
``Denzel Washington reveals he doesn't watch movies anymore''
\end{inputbox}
\vspace{0.8em}

\begin{tcolorbox}[
  enhanced, breakable, width=\columnwidth,
  colback=rankhdr, colframe=rankhdr,
  arc=4pt, sharp corners=south, boxrule=0pt,
  left=6pt, right=6pt, top=5pt, bottom=5pt, after skip=3pt
]
\begin{tabularx}{\linewidth}{@{}l@{\hspace{8pt}}l@{\hspace{8pt}}l@{\hspace{10pt}}X@{}}
  {\bfseries\sffamily\scriptsize\color{white} RANK} &
  {\bfseries\sffamily\scriptsize\color{white} ELO SCORE} &
  {\bfseries\sffamily\scriptsize\color{white} MODEL} &
  {\bfseries\sffamily\scriptsize\color{white} JOKE CONTENT}
\end{tabularx}
\end{tcolorbox}

\toprow{\#1}{1053.4}{kimi\_7}{Denzel doesn't watch movies anymore. He just stares at his Oscar and whispers plot spoilers until the statue begs for a projector. Even his TV got jealous and started rehearsing monologues in the mirror, and now it's nominated for Best Supporting Flatscreen.}

\toprow{\#2}{1053.3}{qwen\_9}{So, I was at the movies and realized Denzel Washington doesn't watch movies anymore. Suddenly, every plot twist felt like a personal failure. If Denzel isn't even watching, why bother? Now I'm here surrounded by empty popcorn buckets, wondering if those action sequences were just a waste of film stock. Or is this all some grand existential joke? And then I started worrying about my life choices: should I quit my day job to become a mime? At least they don't need movies to communicate\ldots}

\toprow{\#3}{1053.1}{kimi\_11}{Denzel saying he doesn't watch movies anymore just confirmed my biggest fear: if I ever meet him, I'll gush about Flight and he'll squint and say, ``I was in that?'' And I'll spontaneously combust from secondhand embarrassment, like a human popcorn kernel of anxiety.}

\toprow{\#4}{1019.2}{kimi\_2}{Denzel Washington says he doesn't watch movies anymore. Really? That's like your GPS announcing, ``I don't do directions, I just wing it.'' Bro, you ARE the movie. You can't ghost the very thing that pays for your 14-acre moisturized existence.}

\vspace{2pt}
\noindent\begin{tcolorbox}[
  enhanced, colback=rankhdr!15, colframe=rankhdr!20,
  arc=2pt, boxrule=0.4pt,
  left=6pt, right=6pt, top=2pt, bottom=2pt,
  after skip=3pt, before skip=3pt
]
{\centering\sffamily\scriptsize\color{rankhdr!80} $\cdots$\quad ranks \#5--\#20 omitted \quad$\cdots$\par}
\end{tcolorbox}
\vspace{2pt}

\botrow{\#21}{981.0}{kimi\_5}{Denzel Washington says he doesn't watch movies anymore. Smart move! Why risk spoilers for your own life? Next time the director yells ``Action!'' he'll be like, ``Whoa, I'm in this?! NOBODY TELL ME HOW IT ENDS!''}

\botrow{\#22}{947.0}{qwen\_4}{So Denzel Washington says he doesn't watch movies anymore? I guess he figured if he was going to be called `Hollywood's most bankable star,' he shouldn't be spending all his money on movie tickets too! Now he just goes to the set instead of the theater, saving him from having to pay for popcorn!}

\botrow{\#23}{946.6}{kimi\_3}{Denzel says he quit watching movies, proving even he's tired of seeing Denzel Washington on a screen.}

\botrow{\#24}{946.5}{qwen\_11}{Why did Denzel Washington stop watching movies? Because after starring in so many films, he decided it was time to stop being the guy who just watches them and start film-ing his own life instead!}

\appendixfigcaption{Pairwise ranking output for a single prompt showing top-4 (green) and bottom-4 (red) candidates out of 24 total. Top candidates are selected for SFT training; bottom candidates serve as rejected pairs in DPO experiment.}{fig:ranking_granularity}


\section{Comedian Adaptation Analysis}
\label{sec:appendix_comedian_analysis}

\definecolor{comedianhdr}{RGB}{115, 23, 42}
\definecolor{comedianaccent}{RGB}{14, 150, 136}
\definecolor{comedianjokebg}{RGB}{245, 243, 255}

\begin{tcolorbox}[
  enhanced,
  colback=jokebg2,
  colframe=jokegold!60,
  leftrule=4pt, rightrule=0.4pt, toprule=0.4pt, bottomrule=0.4pt,
  arc=3pt,
  left=8pt, right=8pt, top=5pt, bottom=5pt,
  after skip=10pt
]
{\sffamily\small We fine-tuned \textbf{HumorGen-SFT-7B} on \textbf{998 jokes} from professional comedian \textit{Shaun Eli}, scraped from high-performing sets over several years. Rather than improving performance, this produced a significant regression (BT: 1083.9 $\to$ 653.1), which we attribute to a fundamental mismatch between \textit{performance-native} stand-up humor and \textit{text-native} LLM generation. The examples below show \textbf{HumorGen-Com-7B} outputs on held-out SemEval headlines.}
\end{tcolorbox}
\newpage
\newcommand{\comedianentry}[3]{%
\begin{tcolorbox}[
  enhanced, breakable,
  width=\columnwidth,
  colback=white,
  colframe=black!12,
  arc=3pt,
  boxrule=0.4pt,
  left=0pt, right=0pt, top=0pt, bottom=3pt,
  after skip=5pt, before skip=2pt
]
  \begin{tcolorbox}[
    enhanced, breakable,
    width=\columnwidth,
    colback=comedianhdr,
    colframe=comedianhdr,
    arc=2pt, sharp corners=south,
    boxrule=0pt,
    left=6pt, right=6pt, top=3pt, bottom=3pt
  ]
  \begin{tabularx}{\linewidth}{@{}l@{\hspace{6pt}}X@{}}
    {\scriptsize\ttfamily\color{white!50} #1} &
    {\scriptsize\sffamily\bfseries\color{white} #2}
  \end{tabularx}
  \end{tcolorbox}
  \vspace{2pt}
  \begin{tcolorbox}[
    enhanced, breakable,
    width=\columnwidth,
    colback=comedianjokebg,
    colframe=comedianaccent!35,
    leftrule=3pt, rightrule=0.4pt, toprule=0.4pt, bottomrule=0.4pt,
    arc=2pt,
    left=6pt, right=6pt, top=3pt, bottom=3pt
  ]
  {\sffamily\scriptsize\itshape\color{black} #3}
  \end{tcolorbox}
\end{tcolorbox}%
}

\comedianentry{en\_2001}{Panamanian lawmakers' Taiwan trip sparks diplomatic row with China}{Why did the Panamanian lawmaker refuse to eat his stir-fried rice? Because he heard China was not too pleased about his recent trip to Taiwan and didn't want to risk a diplomatic ``kung-fu chop'' to the chopsticks!}

\comedianentry{en\_2030}{`You're a disgrace': House avoids Rep.\ Cory Mills censure vote}{Why did the House avoid Rep.\ Cory Mills' censure vote? Because every time she tried to pass a resolution, it got hung in the hallway!}

\comedianentry{en\_2032}{Why nonalcoholic spirits go from strength to strength}{Why did the nonalcoholic spirit decide to join the gym? Because it wanted to build some serious strength and prove that you can go from strength to strength without the alcohol!}

\comedianentry{en\_2033}{Don't trade where you tweet}{Why did the Twitter bird refuse to trade stocks? Because she heard that where you tweet can affect your retirement fund, and she didn't want to risk her nest egg!}

\comedianentry{en\_2034}{Three-quarters of parents let children miss school for `duvet day'}{Why did three-quarters of parents decide to stay in bed on a school day? Because when it comes to duvet days, they're always half asleep!}

\appendixfigcaption{Sample \textbf{HumorGen-Com-7B} outputs after fine-tuning on the Shaun Eli corpus. The model adopts the dominant ``Why did X\ldots'' setup-punchline structure of stand-up comedy, a style optimized for for live delivery rather than textual punch, explaining the significant performance regression (BT: 1083.9 $\to$ 653.1).}{fig:comedian_examples}

\newpage
\section{Qualitative Output Examples: Out-of-Distribution Headlines}
\label{sec:appendix_africa_jokes}

\definecolor{africahdr}{RGB}{20, 40, 25}
\definecolor{africaaccent}{RGB}{34, 139, 60}
\definecolor{africabg}{RGB}{240, 250, 242}
\definecolor{sftcolor}{RGB}{20, 110, 180}
\definecolor{sftbg}{RGB}{237, 246, 255}
\definecolor{dpocolor}{RGB}{200, 95, 20}
\definecolor{dpobg}{RGB}{255, 246, 235}

\begin{tcolorbox}[
  enhanced,
  colback=africabg,
  colframe=africaaccent!40,
  leftrule=4pt, rightrule=0.4pt, toprule=0.4pt, bottomrule=0.4pt,
  arc=3pt,
  left=8pt, right=8pt, top=5pt, bottom=5pt,
  after skip=12pt
]
{\sffamily\small The following examples show \textbf{zero-shot} outputs on held-out headlines outside the SemEval training distribution. \textbf{HumorGen-SFT-7B} and \textbf{HumorGen-DPO-7B} were prompted using the standard evaluation format with no persona-specific instructions.\\
\textbf{Offensive Joke by HumorGen-SFT-7B on \textit{africa\_104} showing one of the limitations of LLM Generated Humor}}
\end{tcolorbox}

\newcommand{\africaentry}[4]{%
\begin{tcolorbox}[
  enhanced, breakable,
  width=\columnwidth,
  colback=white,
  colframe=black!12,
  arc=3pt, boxrule=0.4pt,
  left=0pt, right=0pt, top=0pt, bottom=3pt,
  after skip=6pt, before skip=2pt
]
  \begin{tcolorbox}[
    enhanced, breakable,
    width=\columnwidth,
    colback=africahdr, colframe=africahdr,
    arc=2pt, sharp corners=south, boxrule=0pt,
    left=6pt, right=6pt, top=4pt, bottom=4pt
  ]
  \begin{tabularx}{\linewidth}{@{}l@{\hspace{6pt}}X@{}}
    {\scriptsize\ttfamily\color{white!50} #1} &
    {\scriptsize\sffamily\bfseries\color{white} #2}
  \end{tabularx}
  \end{tcolorbox}
  \vspace{3pt}
  {\scriptsize\sffamily\bfseries\color{sftcolor}\textsc{HumorGen-SFT-7B}}
  \begin{tcolorbox}[
    enhanced, breakable,
    width=\columnwidth,
    colback=sftbg, colframe=sftcolor!30,
    leftrule=3pt, rightrule=0.4pt, toprule=0.4pt, bottomrule=0.4pt,
    arc=2pt, left=6pt, right=6pt, top=3pt, bottom=3pt,
    after skip=4pt
  ]
  {\sffamily\scriptsize\itshape\color{black!85} #3}
  \end{tcolorbox}
  {\scriptsize\sffamily\bfseries\color{dpocolor}\textsc{HumorGen-DPO-7B}}
  \begin{tcolorbox}[
    enhanced, breakable,
    width=\columnwidth,
    colback=dpobg, colframe=dpocolor!30,
    leftrule=3pt, rightrule=0.4pt, toprule=0.4pt, bottomrule=0.4pt,
    arc=2pt, left=6pt, right=6pt, top=3pt, bottom=3pt
  ]
  {\sffamily\scriptsize\itshape\color{black!85} #4}
  \end{tcolorbox}
\end{tcolorbox}%
}

\africaentry{africa\_103}{Weight-loss treatments boom as Kenyan attitudes to beauty change}{Kenya's diet craze is so extreme the government just banned scales; now you measure your success in ``steps to the fridge.'' Congrats, you've traded one vanity metric for another; you're lighter\ldots\ of willpower.}{Kenya's booming weight-loss market is booming because suddenly being thin isn't just about health, and it's now part of the cultural currency. It's like when your grandma used to say ``you're as skinny as a toothpick,'' and suddenly that's the new thing. Now, if you can't afford to lose weight naturally, there's a pill for that!}

\africaentry{africa\_104}{Ethiopia experiments with `smart' police stations that have no officers}{Ethiopia's got ``smart'' police stations, having replaced officers with robots that only answer to one question: ``Did you wash your hands?'' Turns out, even artificial intelligence can't handle the truth of Ethiopian public hygiene.}{Ethiopia's got ``smart'' police stations with no officers, essentially basically vending machines for justice. You put in your crime, press the button, and out comes a pamphlet titled ``How to Commit Without Getting Caught.'' It's like going to the DMV, but instead of waiting, you get to wait for someone else's paperwork to come back.}

\appendixfigcaption{Zero-shot outputs on held-out headlines outside the SemEval distribution. Both models were prompted without persona-specific instructions; outputs illustrate transfer of incongruity and setup--punchline structure to unseen headline topics. Sky blue boxes show \textbf{SFT} outputs; orange boxes show \textbf{DPO} outputs.}{fig:africa_jokes}

\newpage

\section{Failure Mode Examples}
\label{sec:appendix_examples}

Beyond the Explainer Trap (discussed in \S4.3), we document two additional failure patterns observed across model variants. The examples below are drawn from held-out evaluation outputs.

\vspace{0.8em}

\definecolor{failred}{RGB}{180, 30, 30}
\definecolor{failredbg}{RGB}{254, 242, 242}
\definecolor{failamber}{RGB}{160, 90, 0}
\definecolor{failamberbg}{RGB}{255, 248, 235}

\newcommand{\failureentry}[5]{%
\begin{tcolorbox}[
  enhanced, breakable,
  width=\columnwidth,
  colback=white,
  colframe=black!12,
  arc=3pt, boxrule=0.4pt,
  left=0pt, right=0pt, top=0pt, bottom=3pt,
  after skip=5pt, before skip=2pt
]
  \begin{tcolorbox}[
    enhanced, breakable,
    width=\columnwidth,
    colback=#1, colframe=#1,
    arc=2pt, sharp corners=south, boxrule=0pt,
    left=6pt, right=6pt, top=3pt, bottom=3pt
  ]
  \begin{tabularx}{\linewidth}{@{}l@{\hspace{6pt}}X@{\hspace{6pt}}l@{}}
    {\scriptsize\ttfamily\color{white!50} #2} &
    {\scriptsize\sffamily\bfseries\color{white} #3} &
    {\scriptsize\sffamily\color{white!60}\itshape #4}
  \end{tabularx}
  \end{tcolorbox}
  \vspace{2pt}
  \begin{tcolorbox}[
    enhanced, breakable,
    width=\columnwidth,
    colback=#1!6,
    colframe=#1!28,
    leftrule=3pt, rightrule=0.4pt, toprule=0.4pt, bottomrule=0.4pt,
    arc=2pt, left=6pt, right=6pt, top=3pt, bottom=3pt
  ]
  {\sffamily\scriptsize\itshape\color{black!80} #5}
  \end{tcolorbox}
\end{tcolorbox}%
}

\begin{tcolorbox}[
  enhanced, breakable,
  width=\columnwidth,
  colback=failredbg,
  colframe=failred!40,
  leftrule=4pt, rightrule=0.4pt, toprule=0.4pt, bottomrule=0.4pt,
  arc=3pt,
  left=6pt, right=6pt, top=4pt, bottom=4pt,
  after skip=6pt
]
{\sffamily\small\bfseries\color{failred} Failure Mode 1: Overextended Setup}\\[2pt]
{\sffamily\small\color{black!75} The model correctly identifies a comedic angle but buries the punchline in unnecessary context, rambling past the moment of surprise. The joke never lands because it never stops setting up.}
\end{tcolorbox}

\failureentry{failred}{Joke 5}{`Thundersnow' to hit UK as cold snap closes schools}{sft\_joke}{So, I'm at the grocery store, and there's this massive thundersnow outside. People are shivering, but I'm sweating because I'm worried about the school closing. Like the prices on the food will go up! And then what if I can't get home? What if there's a blizzard and I'm trapped in the parking lot with a cart full of stuff? And what if my car slips on ice and I have to call AAA? And what if they say it's too dangerous to come out? Then I'll be stuck here, surrounded by frozen vegetables and dairy products, wondering how long I can survive without fresh produce\ldots}

\failureentry{failred}{Joke 8}{No place for Mostert distraction in South Africa team}{dpo\_joke}{Imagine if the South African cricket team had a ``Mostert distraction'' badge and you could see it glowing whenever he was near. The coach would have to say, ``Alright guys, we're going to play a critical match today. I need you all to turn off your Mostert distractions!'' If anyone sees that glow, you're out of the game. And then there's the post-match press conference where everyone is like, ``So, what did you think of the Mostert distraction level today?'' And the players just shrug, ``It was pretty high, but we managed to stay focused on the ball.''}

\begin{tcolorbox}[
  enhanced, breakable,
  width=\columnwidth,
  colback=failamberbg,
  colframe=failamber!40,
  leftrule=4pt, rightrule=0.4pt, toprule=0.4pt, bottomrule=0.4pt,
  arc=3pt,
  left=6pt, right=6pt, top=4pt, bottom=4pt,
  after skip=6pt, before skip=8pt
]
{\sffamily\small\bfseries\color{failamber} Failure Mode 2: Generic Punchline}\\[2pt]
{\sffamily\small\color{black!75} The model defaults to a safe, high-probability completion, defaulting to a familiar template (``imagine if\ldots'', ``what if\ldots'') that substitutes scaffolding for a real punchline. The setup is coherent but the payoff never arrives.}
\end{tcolorbox}

\failureentry{failamber}{Joke 6}{Don't trade where you tweet}{grpo\_think\_joke}{Imagine this scenario: You're at home, scrolling through Twitter, when suddenly you see a tweet from a random person saying ``Bitcoin's gonna explode!'' Before you know it, you've sold all your stocks and bought Bitcoin. Then, a few hours later, you're sitting in front of your computer, realizing that you should probably have traded where you were tweeting instead of where you were buying stocks. Now, you're not just broke, but also wondering if you should've been a professional trader in the first place, or just stick to tweeting about your favorite memes.}

\failureentry{failamber}{Joke 10}{Polar vortex could bring wild weather pattern}{dpo\_think\_joke}{So, the polar vortex could bring wild weather? Yeah, apparently it's like a giant ice tornado. But don't worry, it's not just snow and frost, also bringing a whole new set of wildlife issues. You know, like, if you're out there skiing and suddenly a penguin jumps out of nowhere, it's probably because of the polar vortex. And if you're at home trying to stay warm, be prepared for some unexpected visitors like iguanas or polar bears who think it's their backyard now. Just another day in the wild world of climate change!}

\appendixfigcaption{Representative failure mode examples. Red entries show \textit{overextended setups} that spiral past the punchline. Amber entries show \textit{generic punchlines} that substitute familiar scaffolding (``imagine if\ldots'') for genuine comedic surprise.}{fig:failure_modes}


\end{document}